%% file: main.tex
\documentclass{article}

\usepackage{arxiv}

\usepackage[utf8]{inputenc} 
\usepackage[T1]{fontenc}    
\usepackage{hyperref}       
\usepackage{url}            
\usepackage{booktabs}       
\usepackage{amsfonts}       
\usepackage{amsmath}
\usepackage{nicefrac}       
\usepackage{microtype}      
\usepackage{lipsum}
\usepackage{authblk}
\usepackage[round]{natbib}
\usepackage{graphicx}
\usepackage{subcaption}
\usepackage[ruled,vlined]{algorithm2e}

\DeclareMathOperator*{\argmax}{arg\,max}
\DeclareMathOperator*{\argmin}{arg\,min}

\title{Analyzing the Hidden Activations of Deep Policy Networks: Why Representation Matters}

\author{\textbf{Trevor McInroe}} 
\author{\textbf{Michael Spurrier}}
\author{\textbf{Jennifer Sieber}}
\author{\textbf{Stephen Conneely}}
\affil{Northwestern University}
\affil{\textit {\{trevormcinroe2022, michaelspurrier2021, jennifersieber2022, stephenconneely2020\}@u.northwestern.edu}}

\begin{document}
\maketitle

\begin{abstract}
We analyze the hidden activations of neural network policies of deep reinforcement learning (RL) agents and show, empirically, that it’s possible to know \textit{a-priori} if a state representation will lend itself to fast learning. RL agents in high-dimensional states have two main learning burdens: (1) to learn an action-selection policy and (2) to learn to discern between useful and non-useful information in a given state. By learning a latent representation of these high-dimensional states with an auxiliary model, the latter burden is effectively removed, thereby leading to accelerated training progress. We examine this phenomenon across tasks in the PyBullet Kuka environment, where an agent must learn to control a robotic gripper to pick up an object. Our analysis reveals how neural network policies learn to organize their internal representation of the state space throughout training. The results from this analysis provide three main insights into how deep RL agents learn. First, a well-organized internal representation within the policy network is a prerequisite to learning good action-selection. Second, a poor initial representation can cause an unrecoverable collapse within a policy network. Third, a good initial representation allows an agent's policy network to organize its internal representation even before any training begins.
\end{abstract}

\keywords{Deep reinforcement learning \and Representation learning \and Robotics}
\input{body/intro}
\input{body/background}
\input{body/related}
\input{body/experiments}

\input{body/conclusion}
\newpage

\bibliographystyle{plainnat}  
\bibliography{references.bib}  

\end{document}

%% file: body/intro.tex
\section{Introduction}
The combination of reinforcement learning (RL) and neural networks has gained significant research momentum over the past decade~\citep{rl_hist1, rl_hist2}. The highly expressive nonlinear attributes of neural networks make them a prime candidate for approximating complex functions that can be used as control policies for self-learning agents. This combination has allowed RL algorithms to surpass human performance on complex tasks, such as mastering a suite of classic Atari video games~\citep{dqn} and the board game Go~\citep{alphago}.

Applying deep RL to real-world systems, such as robotics, is difficult due to several fundamental challenges~\citep{kober_bagnell_peters_2013, efficiency}. Perhaps the most apparent is the need for an abundance of agent-environment interactions. This high threshold is highlighted in real-world scenarios due to the relatively slow movement of real systems when compared to digitally-produced systems. This long wall-clock time motivates the search for general methods that can speed up the learning process of deep RL agents.

Many modern robotic tasks are assisted by devices that produce high-dimensional data, such as the images produced by a camera~\citep{robo-camera1, robo-camera2}. Images are rich with information, but there is no guarantee that all of the information present in an image is directly relevant to the task at hand. In certain situations, excess information can essentially act as noise to a machine learning system being trained.

When a deep RL agent is operating in a high-dimensional state space, it has two main learning burdens. For one, it must learn an action-selection policy that achieves the agent's given goal. Second, the neural network policy must learn to discern between useful and non-useful information in the states that it receives. These two objectives are tied together, which creates a noisy learning process that is further exacerbated by RL's well-known high-variance weight updates~\citep{variance}. Ideally, the dimensionality of the state space could be reduced to only the most important axes of variation.

In this paper, we explore the hidden representations within neural network policies that are learning robotic control. Specifically, we examine how neural network policies learn to organize their internal representations of the state space throughout the RL training process. We compare the progress of agents that are trained directly on high-dimensional images from the state space to the progress of agents that are trained on a latent representation that is learned independently from the RL task. Our analysis shows that learning a well-organized internal representation is a prerequisite burden to good policy learning. It also reveals that pre-learned latent representations can significantly improve policy learning by removing this burden. Perhaps surprisingly, our results also show that a well-learned latent representation can give neural network policies an innate internal representation that is well organized even before policy learning has begun.

%% file: body/background.tex
\section{Background}
\subsection{Reinforcement Learning}
We consider the usual reinforcement learning paradigm of a Markov decision process parameterized by the tuple $(\mathcal{A}, \mathcal{S}, \gamma, \mathcal{R})$. In a sequence of discrete time steps, an agent uses a policy $\pi$ that maps states $s \in \mathcal{S}$ to actions $a \in \mathcal{A}$ $\pi : \mathcal{S} \rightarrow \mathcal{A}$. The agent uses scalar rewards $r$ produced by a reward function $\mathcal{R} : \mathcal{S} \times \mathcal{A} \rightarrow \mathbb{R}$ as a feedback mechanism to guide policy improvement. From an optimization perspective, the agent's goal is to learn the optimal policy $\pi^{*}$ that maximizes the long-term expected return $\pi^{*} = \argmax_{\pi \in \Pi} \mathbb{E}_{a \sim \pi}[\sum_{i=t}^{T}\gamma^{i-t}R(s_t,a_t) |s_t=s, a_t=a,\pi]$ where $T$ is the time-horizon of the task, $\gamma \in [0,1)$ is a discount rate that helps the agent weigh between short- and long-term gain, and $\Pi$ is a set of possible policies.

One common family of RL algorithms is \textit{policy gradient methods}. These are approaches that look to optimize $\pi$ directly by computing the gradient of a performance measure with respect to the parameters $\theta$ of the policy. Actor-critic methods are a subclass of policy gradient methods that parameterize both a stochastic policy $\pi(a|s)$ (actor) and a value function $V(s)$ (critic). The actor produces a probability distribution over actions that is conditioned upon the current state. The critic produces a single value meant to represent the understood value of the current state. There are many variations of actor-critic algorithms, and in the remainder of this subsection, we describe the flavor that we implement for this study. In some cases, the actor and the critic are functions that are parameterized independently. In our case, our actor and critic share weights by using a neural network with two prediction heads, one for the actor and one for the critic. For this study, we use the so-called Monte Carlo approach to RL, where the agent accumulates a buffer of states, actions, and rewards during an episode, and then performs a weight update process once the episode has completed.
\begin{algorithm}[!h]
Input: a neural network policy/value function with initialized parameters $\theta$, learning rate $\alpha$\\
\SetAlgoLined
\ForEach{Episode in training}{
Generate episode $\{s_0, a_0, r_1, ..., s_{T-1}, a_{T-1}, r_T \}$ following $\pi_{\theta}$ \\
\ForEach{Step $t$ in the episode}{
$\delta \leftarrow \sum_{i=t}^{T}\gamma^{i-t}R(s_t,a_t) - V_{\theta}(s_t)$ \\
$\theta \leftarrow \theta + \alpha \delta \nabla_{\theta}log \pi_{\theta}(a_t|s_t)$ \\
    }
}
\caption{Monte Carlo Actor-Critic with Shared Weights}
\end{algorithm}

\subsection{Representation Learning}
Representation learning deals with the learning of a function $\Phi$ that can  map data from one dimension to another $\Phi: \mathbb{R}^{d_1} \rightarrow \mathbb{R}^{d_2}$. It can reveal information in data that is not obvious in its original form. As such,  representation learning can be used for analysis purposes, such as converting high-dimensional data into a dimension that can be visualized. Also, it can be used to transform data to improve the performance of downstream modeling tasks. One example of this would be the classic ``kernel trick'', where a toy dataset is made linearly separable by $\mathbb{R}^2 \rightarrow \mathbb{R}^3$.

\subsubsection{Variational autoencoders}
Variational autoencoders (VAEs) are introduced by~\citet{vae} as a self-supervised Bayesian method for learning latent representations of input data. Most frequently, autoencoders use a dual neural network architecture that compresses input data into a lower-dimensional space and then expands that representation back into the input data's original resolution. Specifically, using an encoder $\Phi(z|s)$, we probabilistically produce latent codes $z \in \mathbb{R}^{d_1}$ given a state input $s \in \mathbb{R}^{d_2}$ such that $d_1 < d_2$. In this work, the latent code is drawn from a diagonal Gaussian $z \sim N(\mu, \Sigma)$ where $\Phi$ produces the mean vector $\mu$ and the diagonal $\sigma$ of the covariance matrix $\Sigma$. Finally, a decoder $\Psi(\hat{s}|z)$ takes the latent code and produces an output of the same dimensions as $s$.

As outlined in the original paper, sampling the Gaussian creates a random node in the network that is non-differentiable. We employ a simple solution, the ``reparameterization trick,''  to divert the randomness outside of the main computation graph. This involves using a scaling factor $\epsilon$ that is drawn from a prior $\epsilon \sim p$. In this work, we sample from a unit Gaussian $\epsilon \sim N(0,1)$ and re-compute the latent code $z = \sigma \odot \epsilon + \mu$.

These dual networks can be jointly trained by minimizing some difference between their input and output over a given training set $\mathcal{X}$. The best way to $\argmin_{\Phi, \Psi}\mathbb{E}[\mathcal{X} - \Psi(\Phi(\mathcal{X}))]$ is to have the learned encoder imprint the most important axes of variation from $s$ into $z$. Once the autoencoder is trained, $\Psi$ can be discarded, and we are left with a reliable way to decrease the dimensionality of incoming data while still retaining important information.

\subsubsection{Principal component analysis}
Principal Component Analysis (PCA)~\citep{pca} is a well-known technique for finding a linear transformation to project high-dimensional data down into lower dimensions. These transformed variables, called principal components (PCs), lose information in terms of the raw numbers that represent them, but they largely retain a meaningful distance between data points. As such, PCA is a common method across many scientific disciplines for visualizing and analyzing high-dimensional data~\citep{pca_ex1, pca_exp2, pca_exp3}. PCA works by iteratively finding projections that minimize the distance between the original representation and the PCs while maintaining an orthogonal relationship between each created PC.

%% file: body/related.tex
\section{Related Works}
\subsection{Deep reinforcement learning}
In the last ten years, research has increasingly focused on the application of expressive nonlinear functions, such as neural networks, to reinforcement learning problems~\citep{alphago, alphazero}. The modern-day seminal application in this area was to the domain of video games, accomplished by learning directly on the screen pixels~\citep{dqn}. This work was later expanded upon with various training-stabilization mechanisms~\citep{double, ddpg} as well as enhancements to memory processes that significantly improve sample efficiency~\citep{per}.

In recent years, deep reinforcement learning has successfully been applied to some real-world systems, such as robotics~\citep{meta, pms, emerge, agile}. The applications are diverse, ranging from using human-robot interaction for intrinsically learning social skills~\citep{robots3} to improving classic model predictive control systems via autonomously learning well-performing parameters of a differential drive robot~\citep{robots2}.

\subsection{Representation learning for downstream tasks}
Successfully training deep neural networks is a difficult task. Recently, researchers have looked to use representation learning for the purpose of improving the performance of downstream models. For example,~\citet{patients} apply correlational neural networks~\citep{correl} to learn a latent-space representation of intensive-care patient data to help predict mortality rate. Many modern applications in natural language processing use latent projections of incoming text data produced by models like word2vec~\citep{word2vec}, doc2vec~\citep{doc2vec}, and GloVe~\citep{glove}.

This effort extends into the realm of reinforcement learning.~\citet{atari_stdim} provide a method for learning representations via loss functions that operate directly on the hidden layers of encoder neural networks. They do so by maximizing a mutual information objective~\citep{mutual1, mutual2} across temporally-similar images in Atari games.~\citet{curl} offer an alternative method by using contrastive learning via a dual-encoder network that learns to discriminate between augmented pairs of similar and dissimilar images.

\subsection{Explainable artificial intelligence}
The field of ``explainable artificial intelligence'' seeks to understand \textit{how} and \textit{why} neural networks learn. One early example of this effort is given by~\citet{explainprime}, wherein the authors examine the magnitude of hidden activation values to determine that a well-trained Deep Belief Network learns to detect corners of objects. ~\citet{explain0} quickly built upon this work, proposing several, more automated, methods that establish intuitions about the hierarchical way in which neural networks learn on images. More recently,~\citet{explain1} introduce the global average pooling operation for convolutional neural networks in combination with the class activation mapping technique. This approach gives visual insight into localized areas of an input image that help the network determine classification.

%% file: body/experiments.tex
\section{Experiments}
\subsection{Environment}
We train our agents within the Kuka environment provided by the PyBullet physics engine. The Kuka environment contains a robot gripper arm, a tray, and several objects. The agent's goal is to control the gripper arm to grasp and lift the object out of the tray. The reward function returns a small, negative value that approaches zero as the gripper moves closer to the object and a large, positive value for when the task is successfully completed. To discretize the action-space, we use the following movement matrix:
\begin{equation*}
\begin{bmatrix}
0 & -0.005 & 0.005 & 0 & 0 & 0 & 0 \\
0 & 0 & 0 & -0.005 & 0.005 & 0 & 0 \\
0 & 0 & 0 & 0 & 0 & -0.005 & 0.005
\end{bmatrix}
\end{equation*}
where the rows correspond to movement along the $x$, $y$, and $z$ axes, respectively. The agent's chosen action $a$ extracts a column vector from this matrix. For example, when $a=2$, the movement vector is $<-0.005, 0, 0>$. Once selected, the action repeats 25 times within the environment, and the length of each episode is 40 selected actions. At each time step, the environment produces an image via a static camera. For an example of the state, see Figure~\ref{fig:vae-training}, below.

\subsection{Policy Models and VAEs}
The agents in this work operate in two distinct state spaces. One set of agents learns directly on the raw pixels that are produced by the environment. The other set of agents learns on latent codes produced by a VAE that is trained before the agents begin to learn. The agents that learn directly on the pixels use a policy that is parameterized by a convolutional neural network. The agents that use the latent codes use a fully-connected neural network policy. The base of the convolutional policies have three convolutional layers, each with a 3x3 kernel and a stride of 2, and one dense layer that outputs a vector $x_{conv} \in \mathbb{R}^{64}$. The fully-connected policies have a base made of two dense layers with 32 and 64 hidden units, respectively, that also outputs a vector $x_{dense} \in \mathbb{R}^{64}$. Each policy type has two prediction heads, one for the actor $\pi_{\theta}(a|s)$ and one for the critic $V_{\theta}(s)$. The critic outputs a scalar that values $s$ and the actor outputs a probability distribution that our agents use to sample actions. Both network types use ReLU nonlinearities~\citep{relu} $max(0, x)$ between each layer.

All agents use a Monte Carlo paradigm where states, actions, and rewards are accumulated during an episode and the parameters $\theta$ are updated once the episode completes. Because our actor and critic share parameters via the base of the policy networks, they are effectively updated together in a single backpropagation step that uses the following loss:
\begin{gather*}
    \mathcal{L}(\theta) = \sum_{n=1}^{N} -log\pi_{\theta}(a_n|s_n)[V_{\theta}(s_n) - \sum_{j=1}^{n}\gamma^{n-j} r_{j}] + \sum_{n=1}^{N} |V_{\theta}(s_n) - \sum_{j=1}^{n}\gamma^{n-j} r_{j}|
\end{gather*}
where $N$ is the number of steps taken in the given episode.

\begin{figure}[!h]
    \centering
    \includegraphics[width=0.75\textwidth]{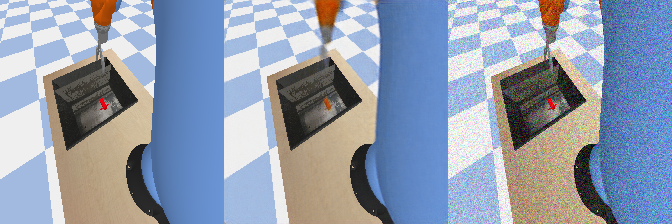}
    \caption{Trio of images produced during VAE training. The original state (left), the image produced by the VAE (center), and the noisy input image (right).}
    \label{fig:vae-training}
\end{figure}

All VAEs are trained on a de-noising task. To do this, we first generate a collection $\mathcal{D}$ of 10,000 images from the environment using an agent that selects actions at random. Then, at each training step, we sample from this collection $d \subseteq \mathcal{D}$ and produce a ``noisy'' version of the images $\bar{s}$ by adding a small amount of Gaussian noise $\sim N(0,0.1)$ to all three color channels of all pixels. These noisy images are used as the VAE's inputs, and the non-noisy versions are used as the VAE's targets. The loss function is a combination of a pixel-wise loss as a reconstruction error, as well as the KL divergence between the encoder $\Phi(\cdot)$ and a prior on the latent variable $Pr(z)$, which is set to $\sim N(0,1)$ .
\begin{gather*}
    \mathcal{L}(\Phi, \Psi) = \frac{1}{M}\sum_{m=1}^{M} \lVert s_m -\Psi(\Phi(\bar{s}_m)) \rVert_{2}^{2} + D_{KL, \bar{s} \in d, z \sim \Phi}[\Phi(z|\bar{s}) \; || \; Pr(z)]
\end{gather*}
where $M$ is the batch size. We train all VAEs with a batch size of 32 for 20,000 epochs before they are deployed into the RL loop. By training the VAE independently of the RL agents, we effectively decouple the learning of a useful representation with the learning of a strong policy. For a visual example of this training process, see Figure~\ref{fig:vae-training}, below.

\subsection{Learning to pick up objects with deep reinforcement learning} 
We deploy all experiments within two distinct tasks. First, with a static object that is placed in the same location every episode (\textit{static, static}). Second, with a static object that is placed at a random point in the tray with a random orientation every episode (\textit{static, random}). 

\begin{figure}[!h]
    \centering
    \begin{subfigure}{\textwidth}
        \centering
        \includegraphics[width=0.45\textwidth]{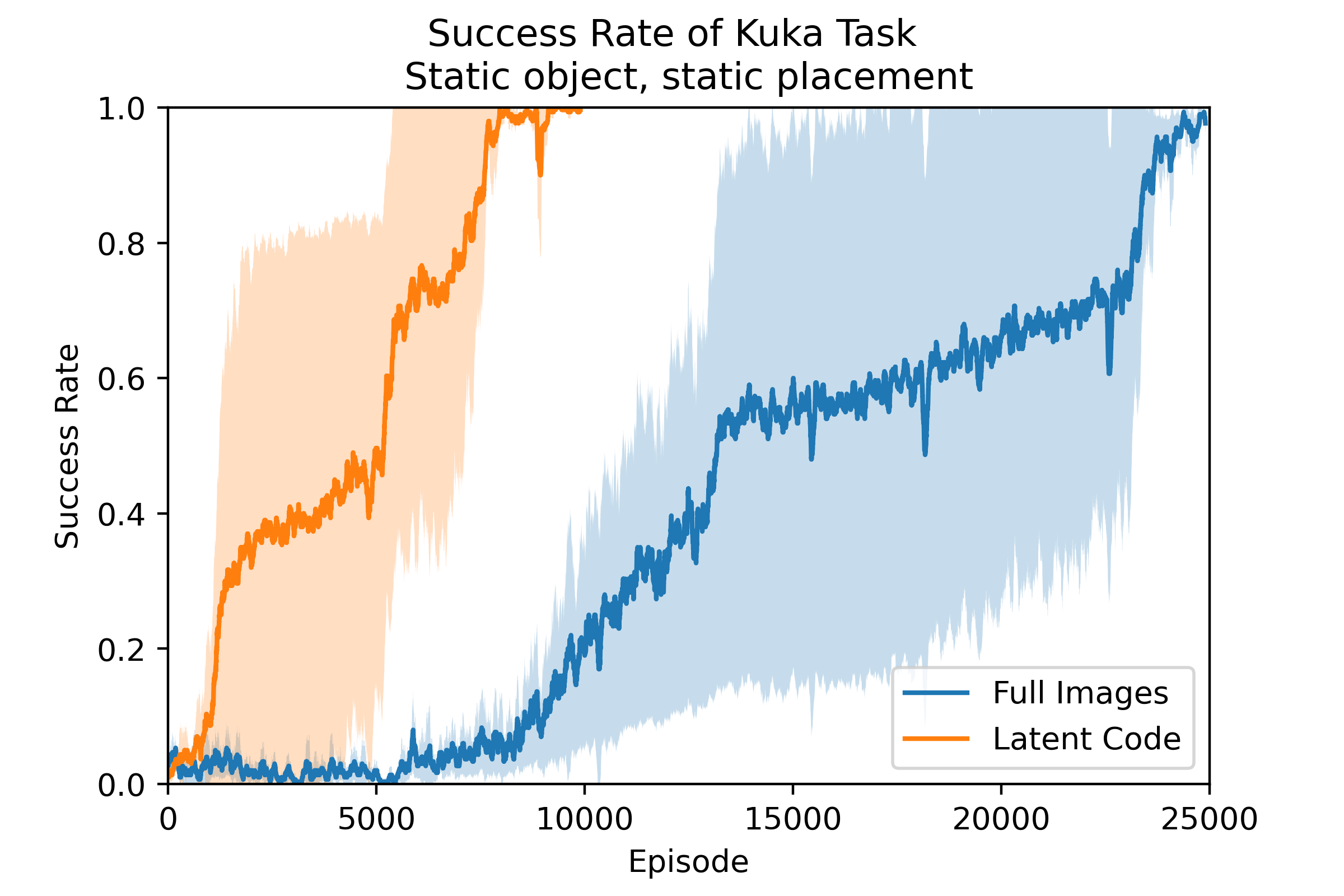}
        \includegraphics[width=0.45\textwidth]{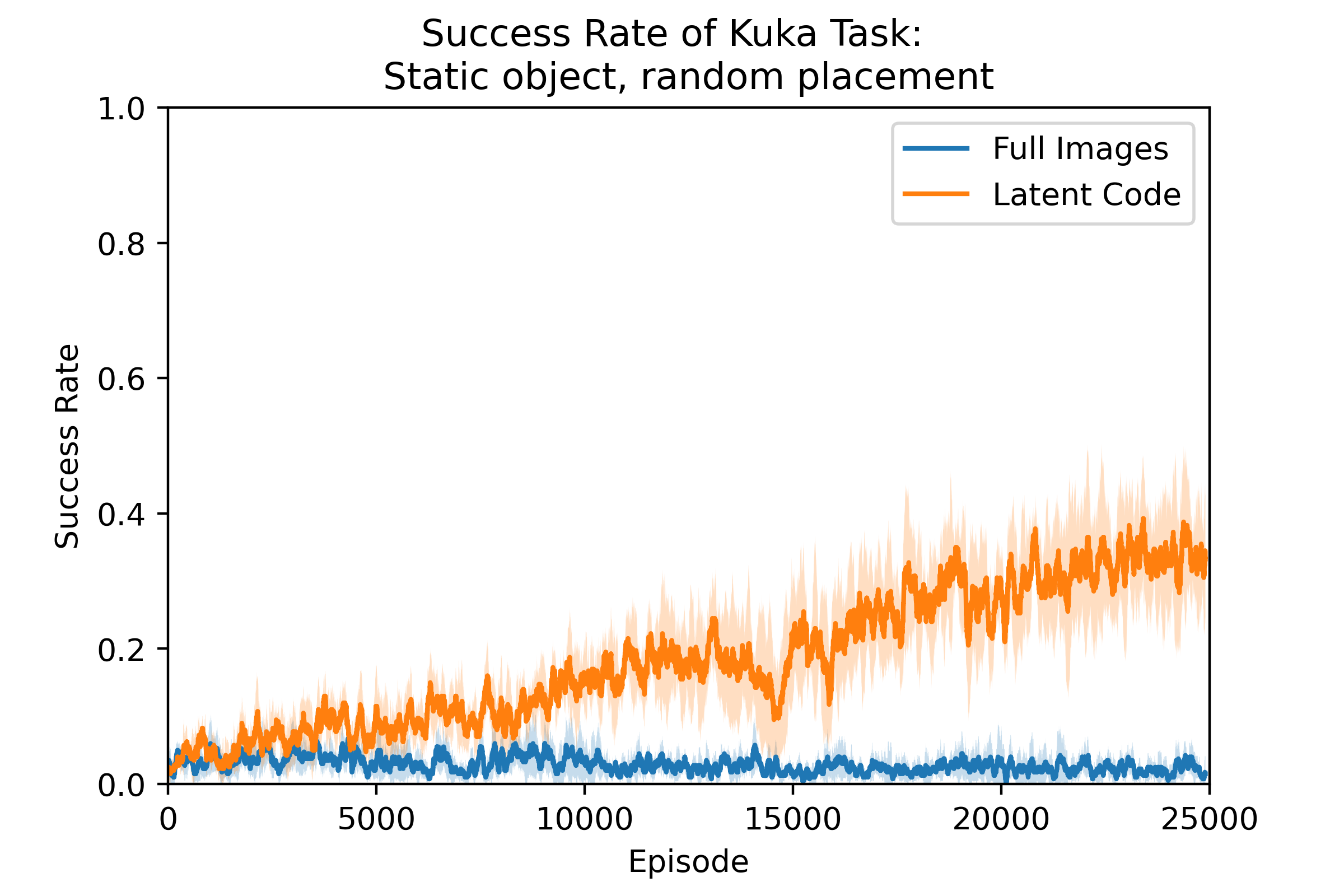}
    \end{subfigure}
    \caption{Success rate throughout training on \textit{static, static} task (left) and \textit{static, random} task (right). Mean performance (bold line) and +/- one standard deviation (shaded area).}
    \label{fig:training-results}
\end{figure}

In each task, each agent is trained three times for roughly 1,000,000 steps each time. The metric for performance is the percentage of the time the agent successfully lifts the object out of the tray within a sliding window of 100 training episodes. Figure~\ref{fig:training-results}, below, shows the mean performance (bold line) as well as +/- one standard deviation (shaded area). We note that, in both tasks, the agents that use latent codes learn at a significantly accelerated rate. For the \textit{static, static} task (left), these agents achieve a consistent, nearly-perfect performance by 10,000 training episodes. At this point, the agents that use the raw pixels are only able to achieve the task about 20\% of the time. In the \textit{static, random} task, the agents that use the full images do not learn to complete the task in any meaningful sense, while the agents that use the latent codes are able to pick up the object nearly 40\% of the time. Also, the progress of the agents that use the latent codes shows an upward trend, suggesting that additional training could improve performance further.

\newpage
\subsection{Visualizing hidden representations within policy networks}
To glimpse into how neural network policies are interpreting their input, we visualize the outputs of their hidden layers. To generate all of the plots in the remainder of this section, we first collect three episodes where the agent successfully picks up the object. Collecting only successful episodes guarantees a wide range of state-reward pairs. For each step in each episode, we capture the outputs of the network bases for both types of agents. Finally, we make these hidden spaces visible by projecting them onto a three dimensional manifold using PCA. While the actual transformed values do not carry inherent meaning by themselves, this projection largely retains a meaningful distance between points.

\begin{figure*}[!h]
    \centering
    \begin{subfigure}[b]{0.85\textwidth}
        \centering
        \includegraphics[width=0.49\linewidth]{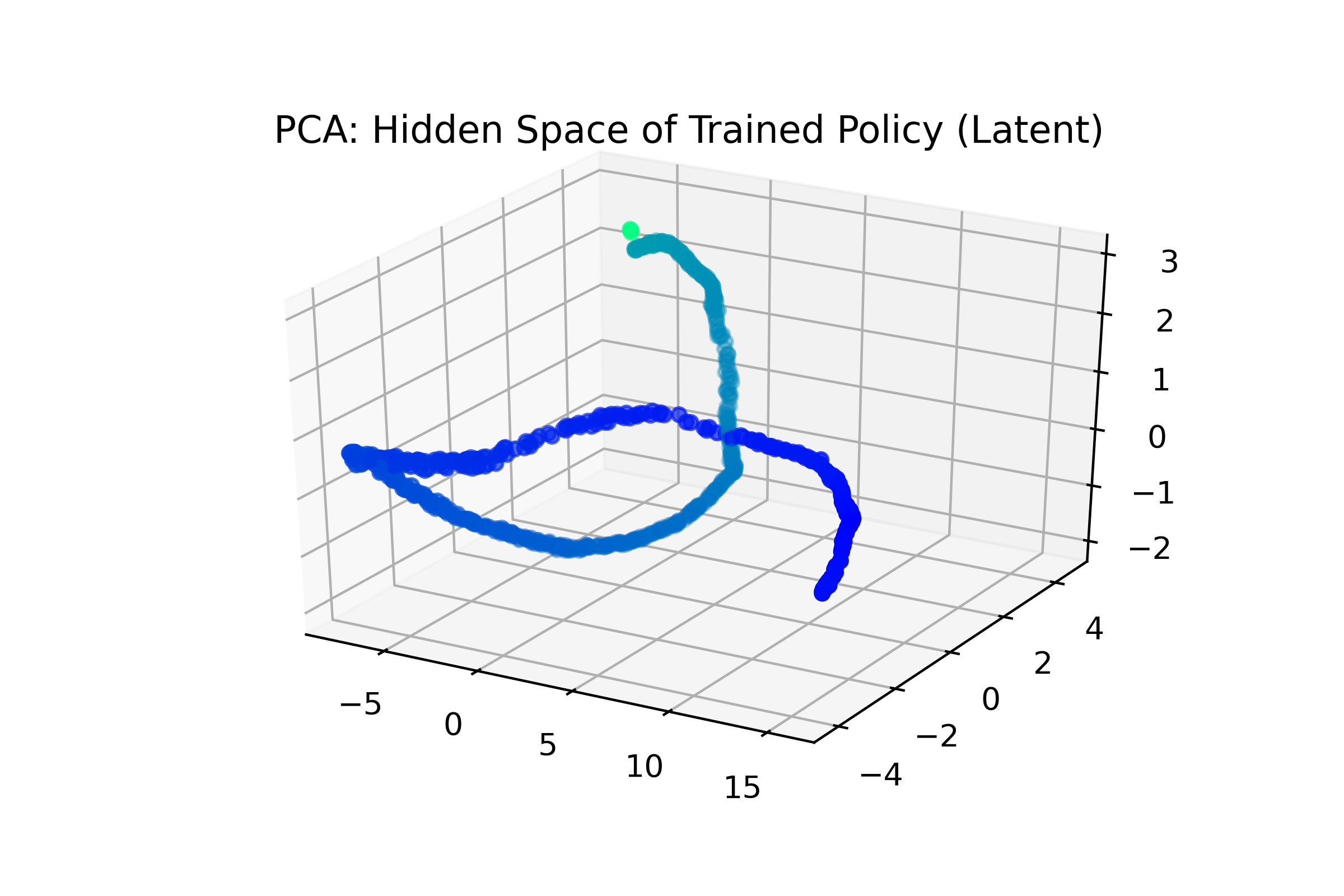}
        \includegraphics[width=0.49\linewidth]{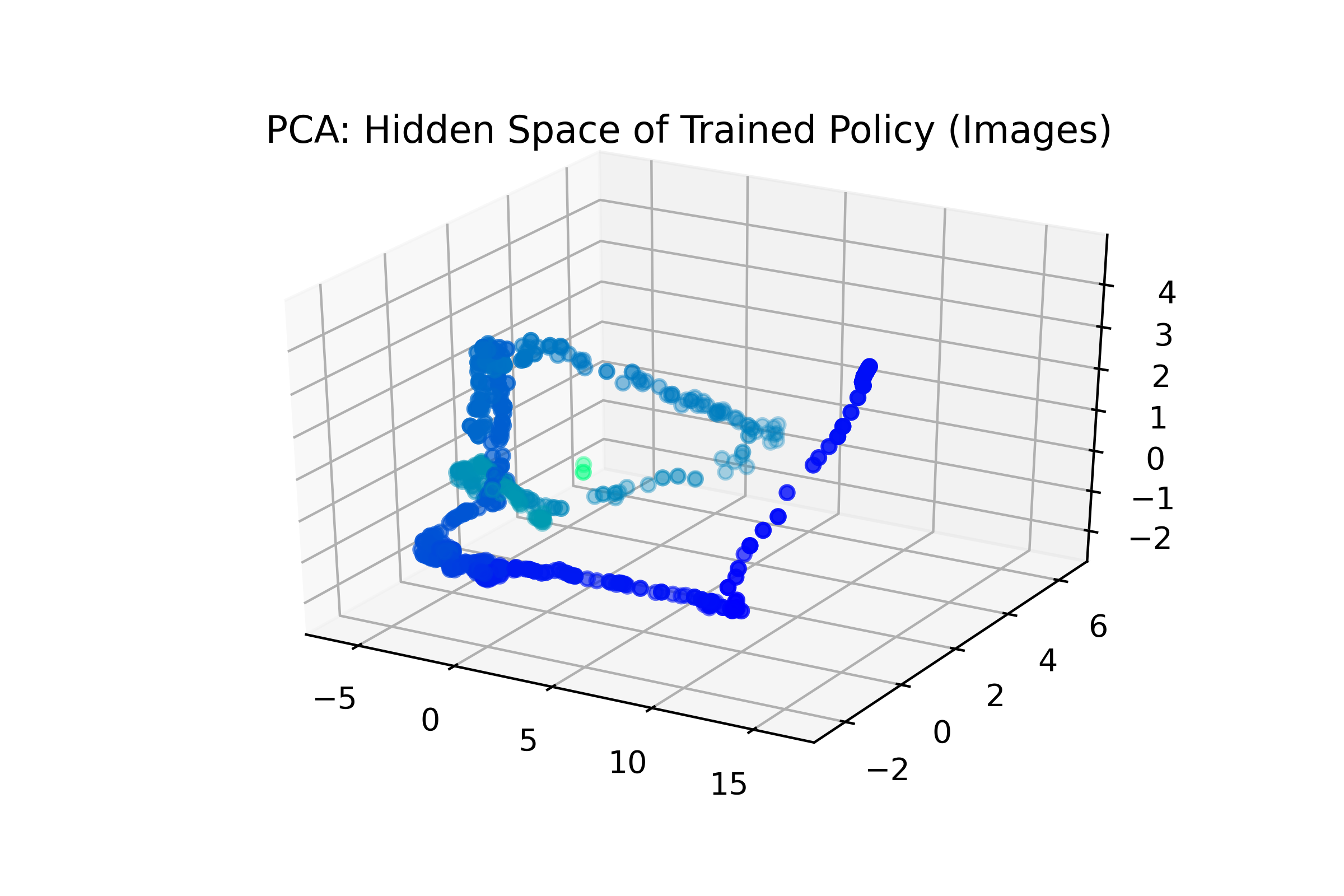}
    \end{subfigure}
    \hfill
    \vskip\baselineskip
    \begin{subfigure}[b]{0.85\textwidth}   
        \centering 
        \includegraphics[width=0.49\linewidth]{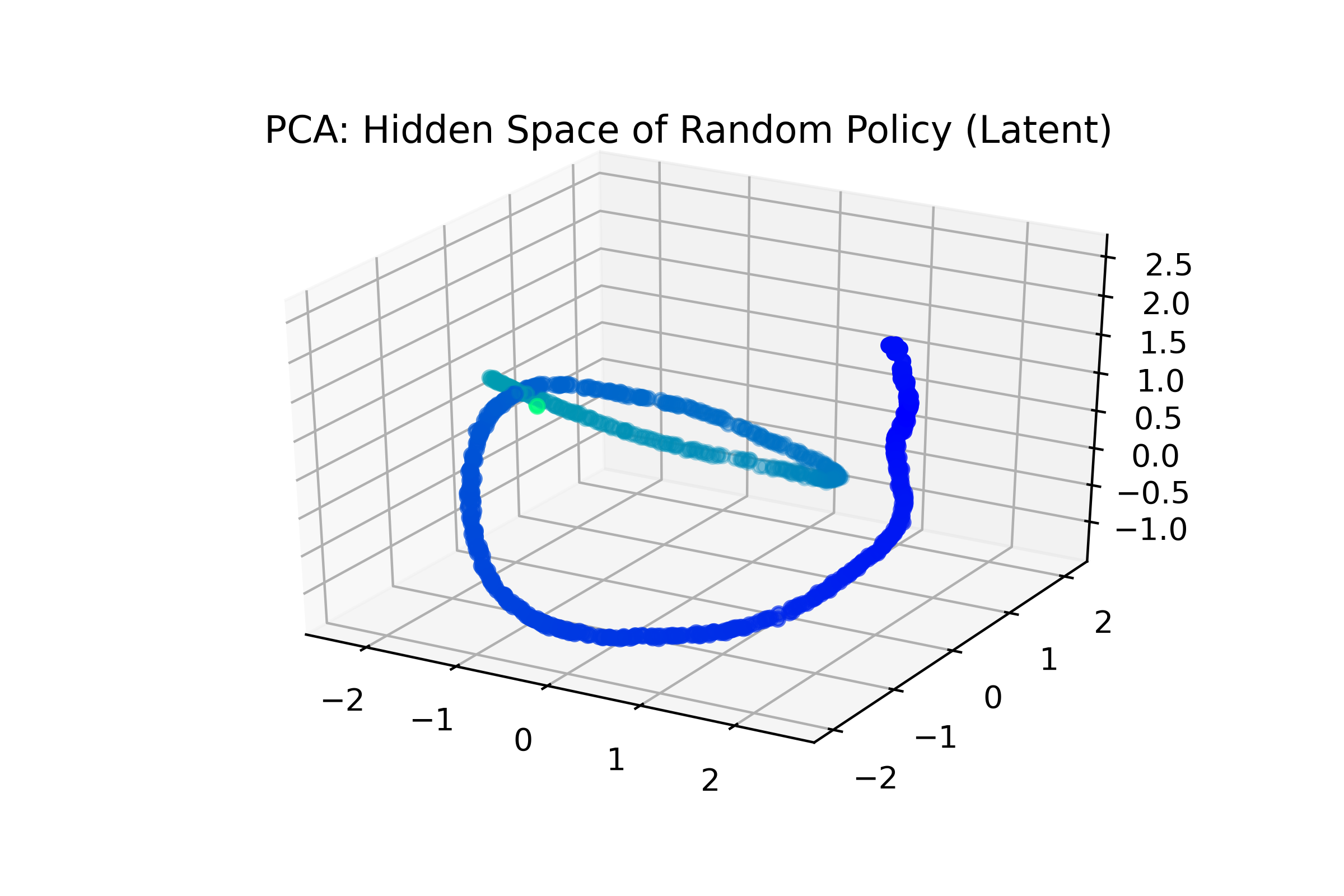}
        \includegraphics[width=0.49\linewidth]{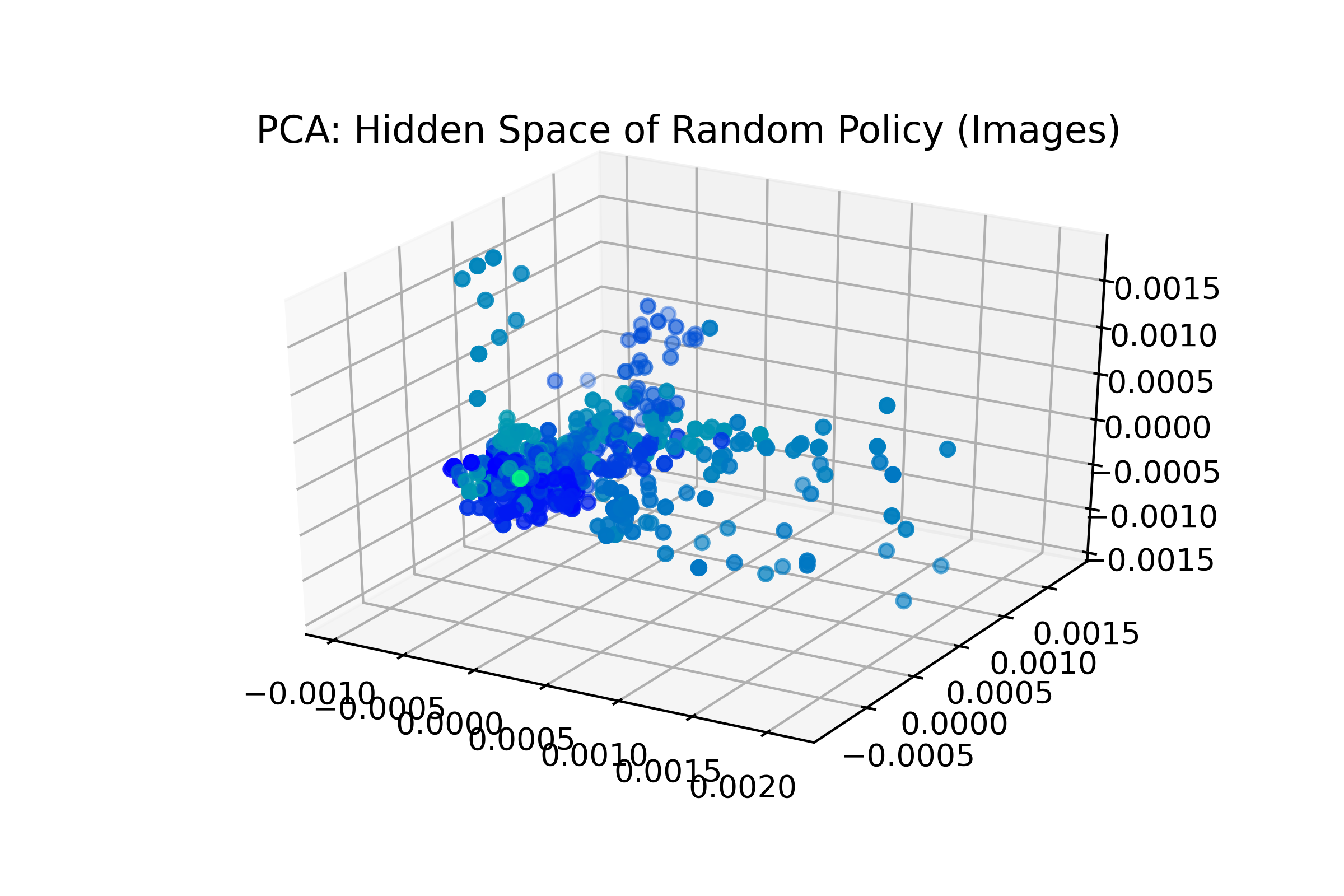}
    \end{subfigure}
    \hfill
    \caption{PCA projections of the hidden spaces of fully-trained latent (top left) and image (top right) policies as well as randomly-initialized latent (bottom left) and image (bottom right) policies.}
    \label{fig:hidden-spaces}
\end{figure*}

Figure~\ref{fig:hidden-spaces}, below, shows these projections for four cases: (a) fully-trained policy that uses latent codes (top left), (b) fully-trained policy that uses the full images (top right), (c) a randomly-initialized policy that uses the latent codes (bottom left), and (d) a randomly-initialized policy that uses the full images (bottom right). The colors of each point correspond to the received reward of that point from lowest (dark) to highest (light). 

We note that both of the fully-trained policies (top row) seem to have learned some structure of the input data. The input states that are close together in terms of their given reward are placed near each other in high-dimensional space by the internals of the neural networks. Surprisingly, the non-trained latent policy network (bottom left) shows a similar level of structure, despite being filled with completely random weights. Further, the non-trained image policy network (bottom right) does not show this. This result suggests that the latent codes, as learned by the VAEs, already carry an explicit structure. 

\begin{figure}[!h]
    \centering
    \begin{subfigure}{0.85\textwidth}
    \centering
        \includegraphics[width=.49\linewidth]{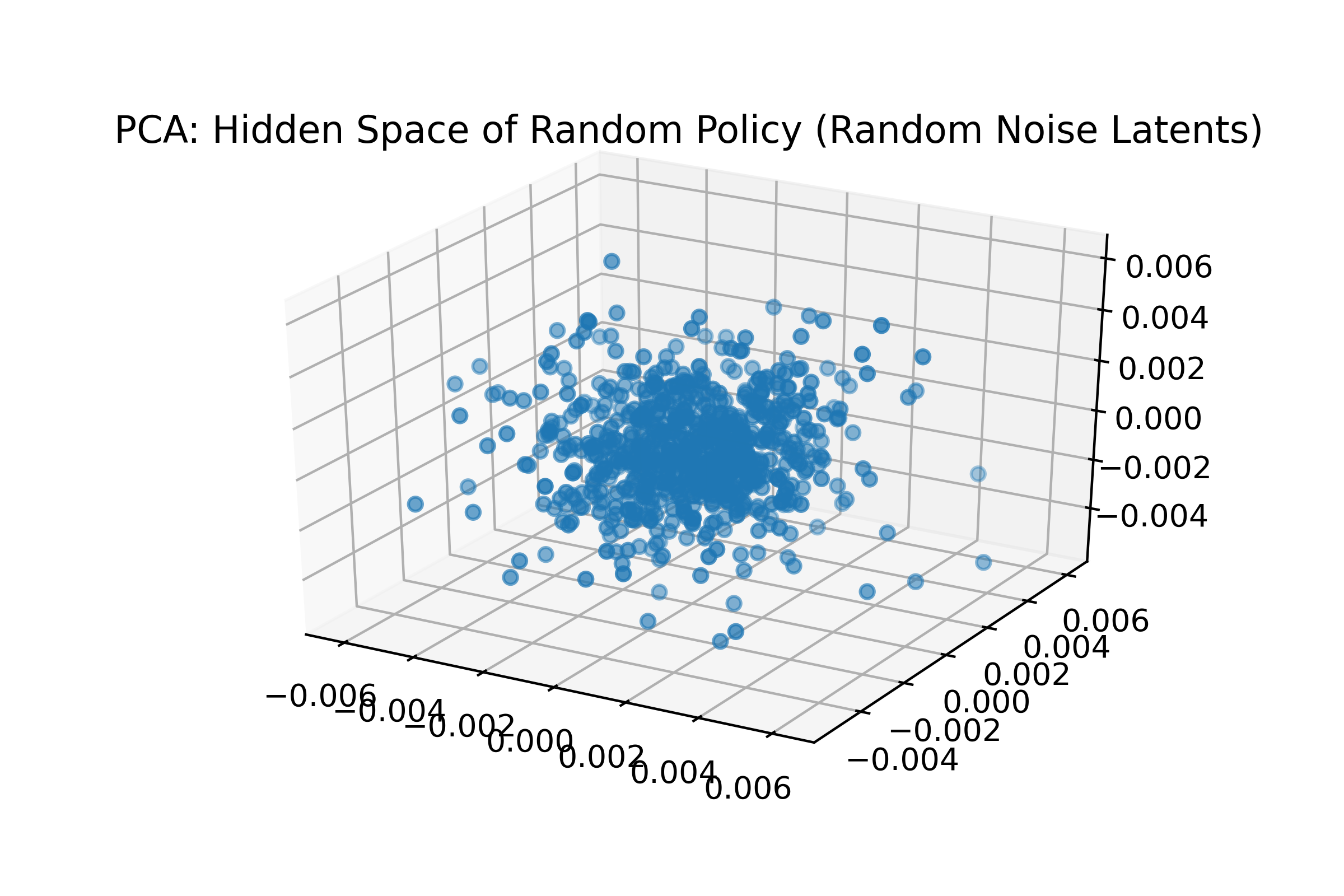}
        \includegraphics[width=.49\linewidth]{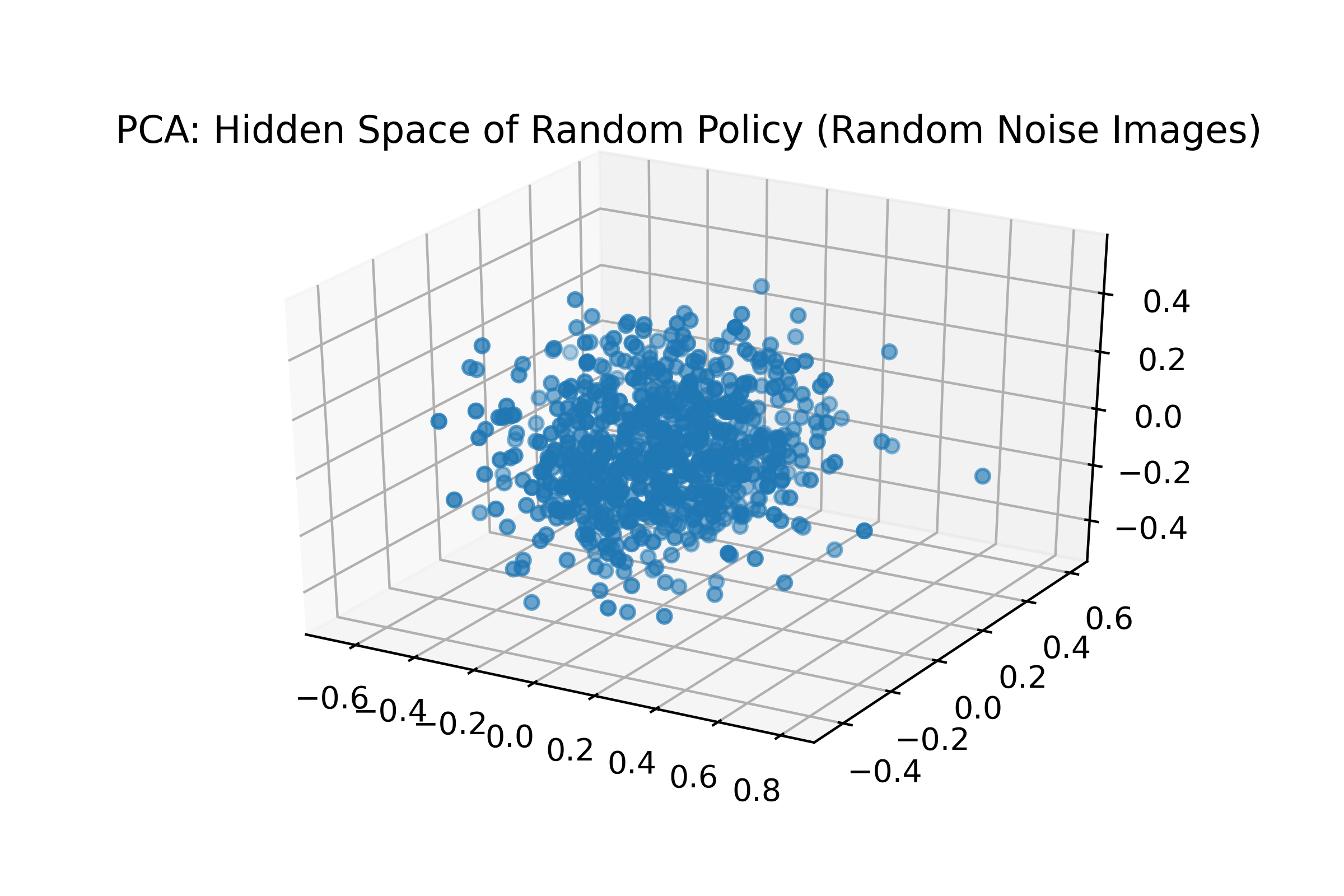}
    \end{subfigure}
    \caption{PCA projection of fully-connected policy (left) and convolutional policy (right) when given random inputs.}
    \label{fig:pca-random}
\end{figure}

As a visual control, Figure~\ref{fig:pca-random} displays the same hidden-space projection when the policies are fed randomly generated data. No color scale is used, as the randomly generated data do not correspond to any state-reward pair. 
\newpage
\subsection{Does weight initialization method affect internal structure?}
\begin{figure*}[!h]
    \centering
    \begin{subfigure}[b]{0.85\textwidth}
        \centering
        \includegraphics[width=0.49\linewidth]{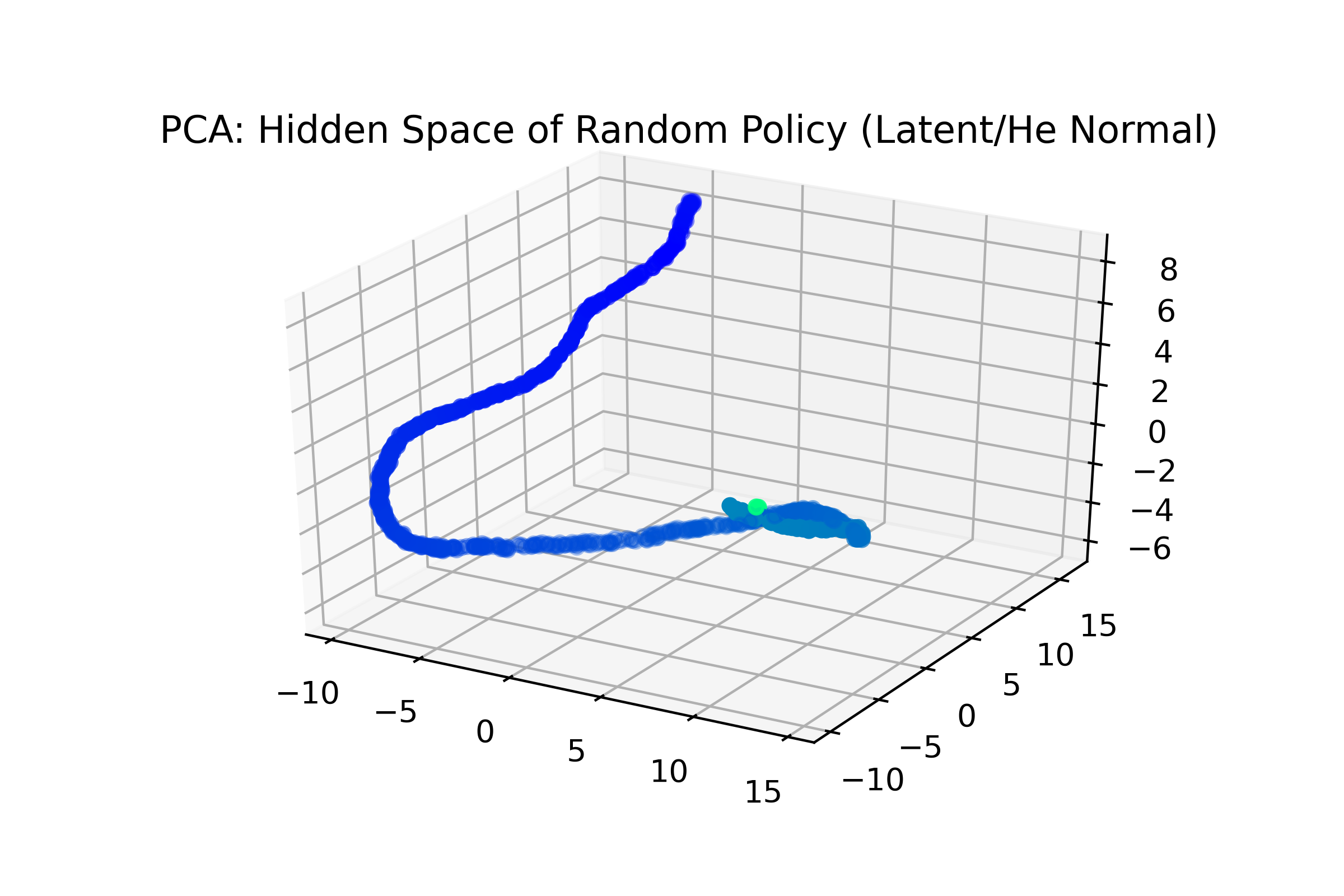}
        \includegraphics[width=0.49\linewidth]{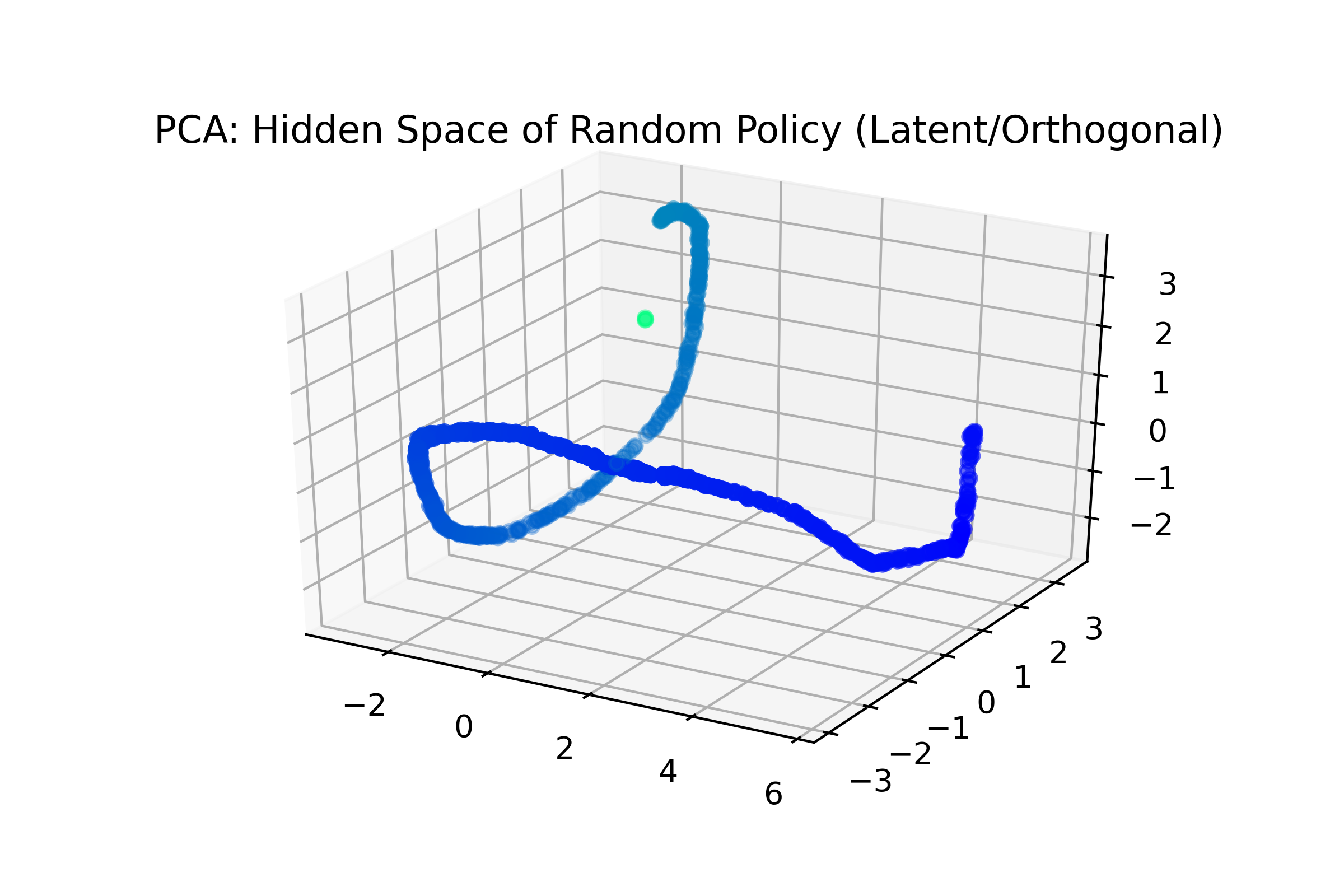}
    \end{subfigure}
    \hfill
    \vskip\baselineskip
    \begin{subfigure}[b]{0.85\textwidth}   
        \centering 
        \includegraphics[width=0.49\linewidth]{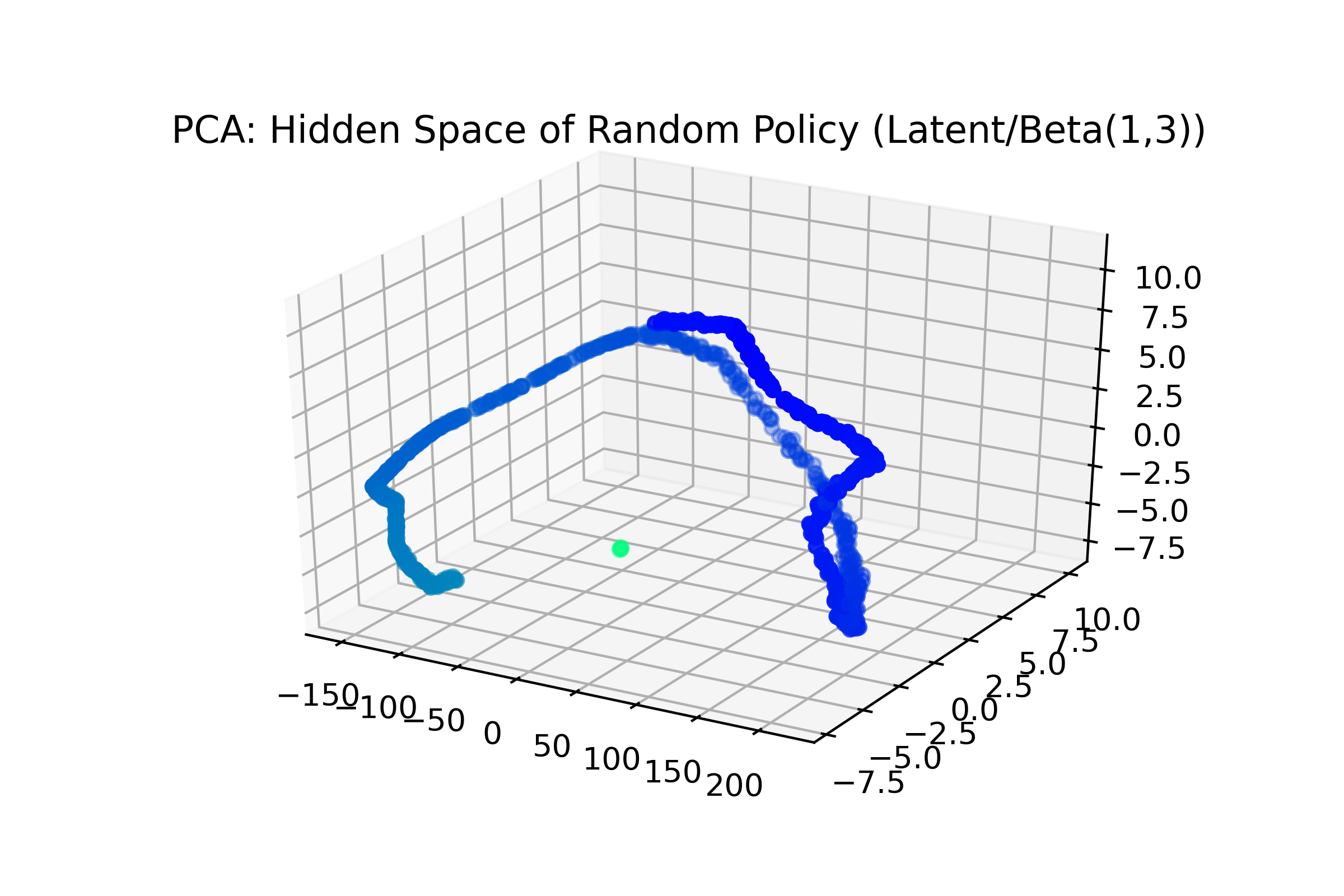}
        \includegraphics[width=0.49\linewidth]{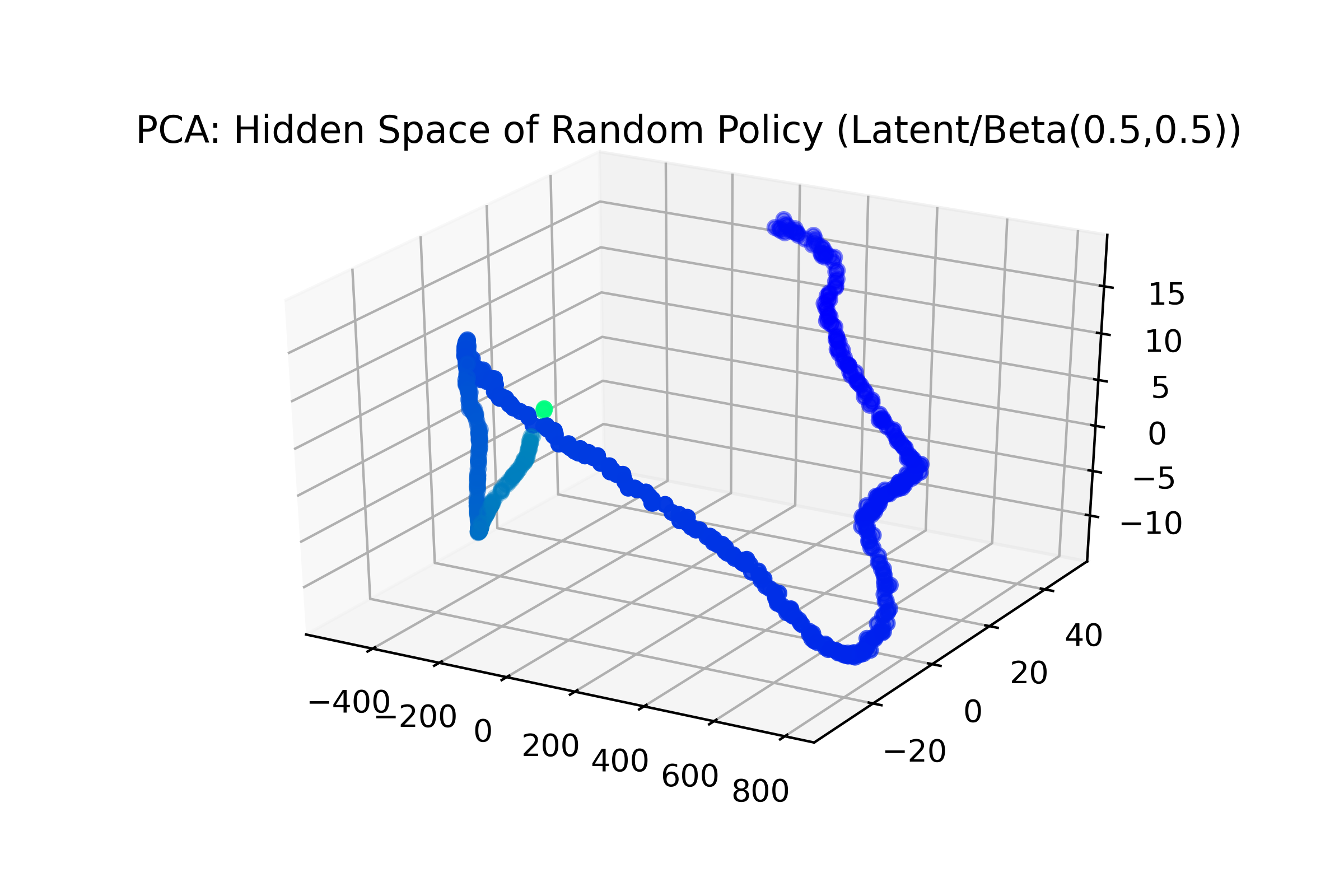}
    \end{subfigure}
    \hfill
    \caption{PCA projection of hidden representations of VAE-learned latent inputs under various weight initialization schemes.}
    \label{fig:latent-inits}
\end{figure*}

To ensure that the results in the previous subsection are not confined to a specific weight initialization scheme, we repeat the PCA projection process on fully-connected policies with two common and two uncommon initialization methods. Figure~\ref{fig:latent-inits}, below, displays networks initialized with He Normal~\citep{he_init} (top left), orthogonal~\citep{ortho} (top right), a one-sided distribution with a long tail ($\sim \beta(1,3)$) (bottom left), and a U-shaped distribution ($\sim \beta(0.5,0.5)$) (bottom right). We note that, across all four weight initialization schemes, each hidden representation retains an organization that relates to reward. These results suggest that the innate internal organization that is gifted by a well-learned latent representation is robust to differences in weight initialization method.

\subsection{How do hidden representations evolve throughout training?}
\begin{figure}[!h]
    \centering
    \begin{subfigure}{\textwidth}
    \centering
        \includegraphics[width=.24\linewidth]{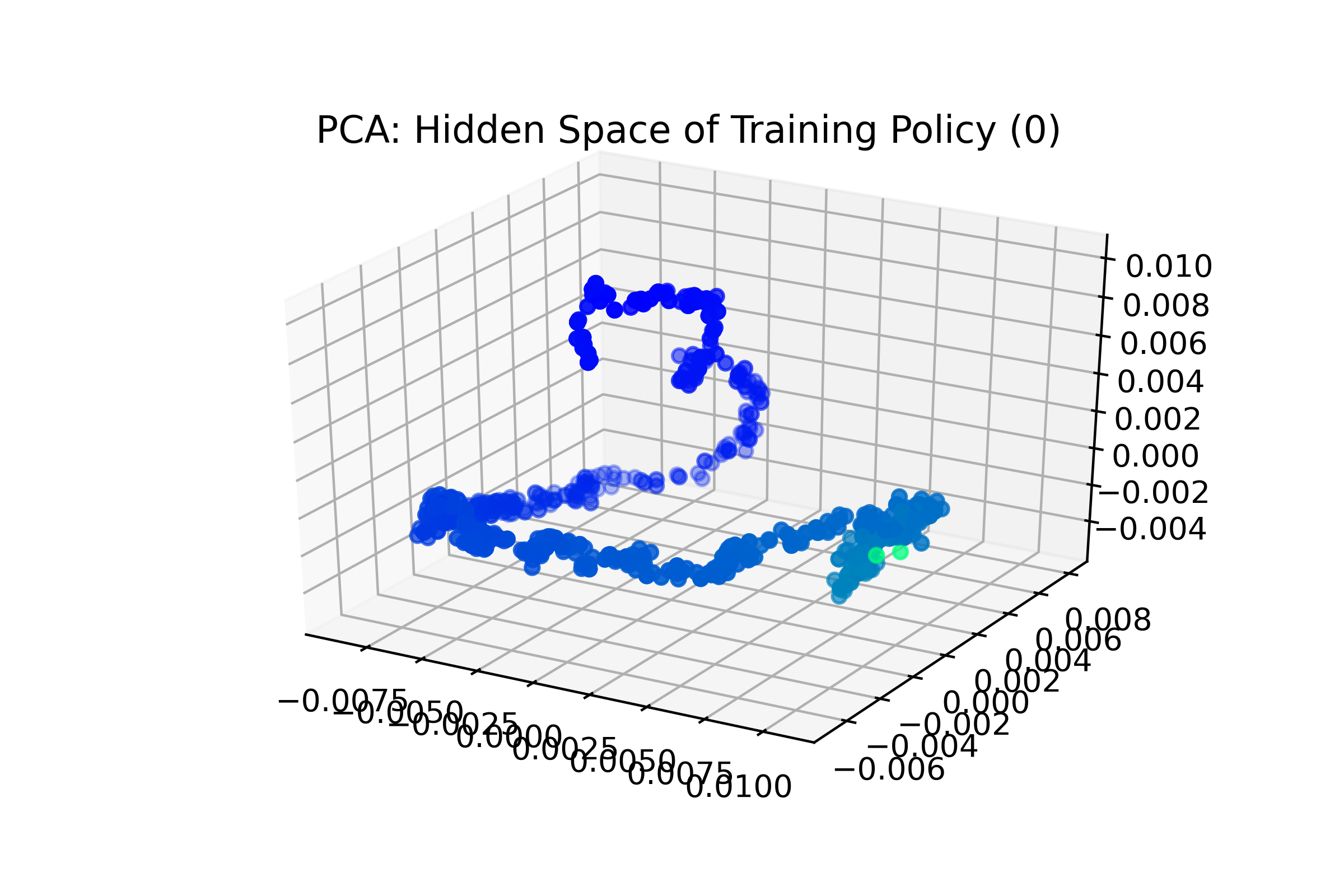}
        \includegraphics[width=.24\linewidth]{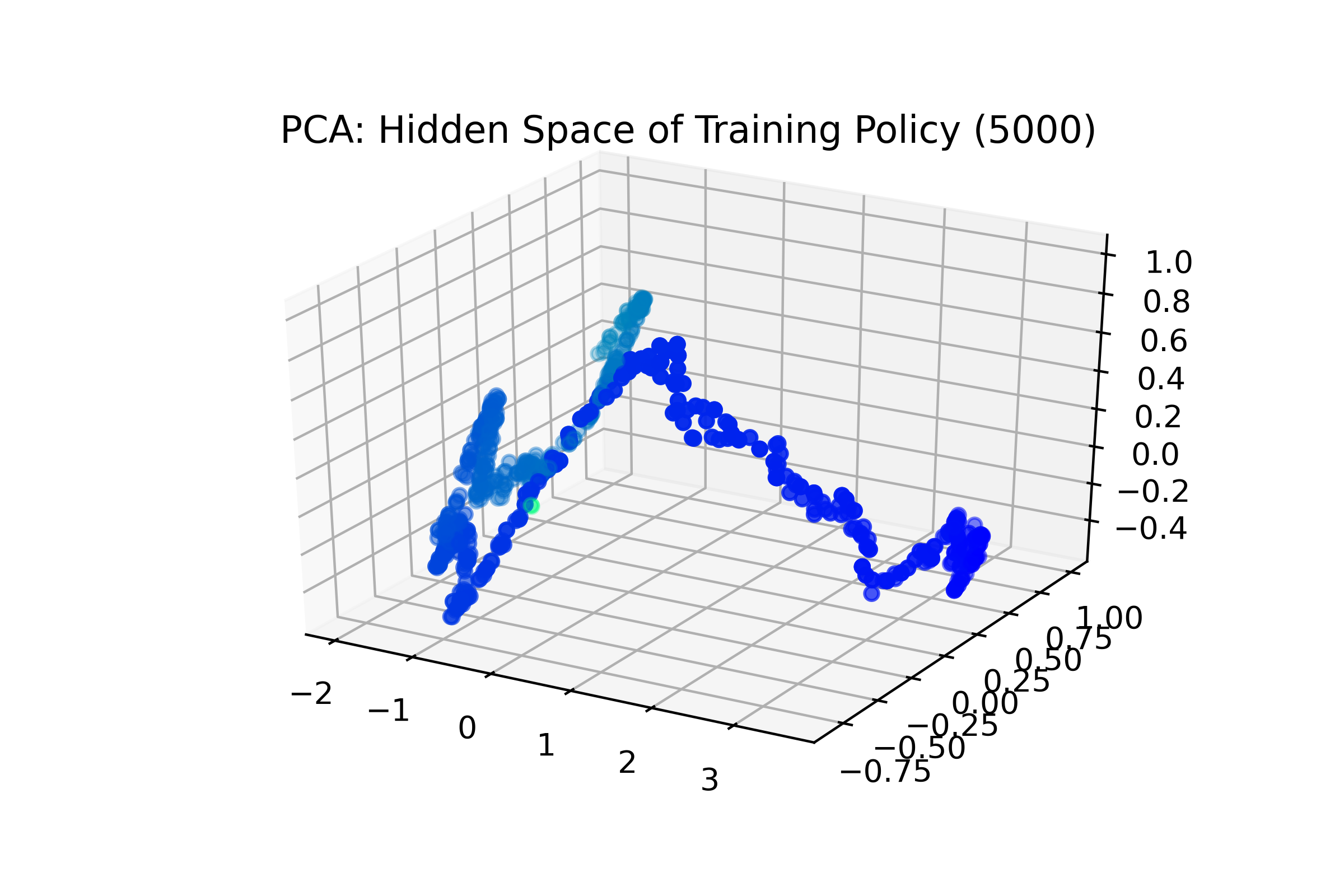}
        \includegraphics[width=.24\linewidth]{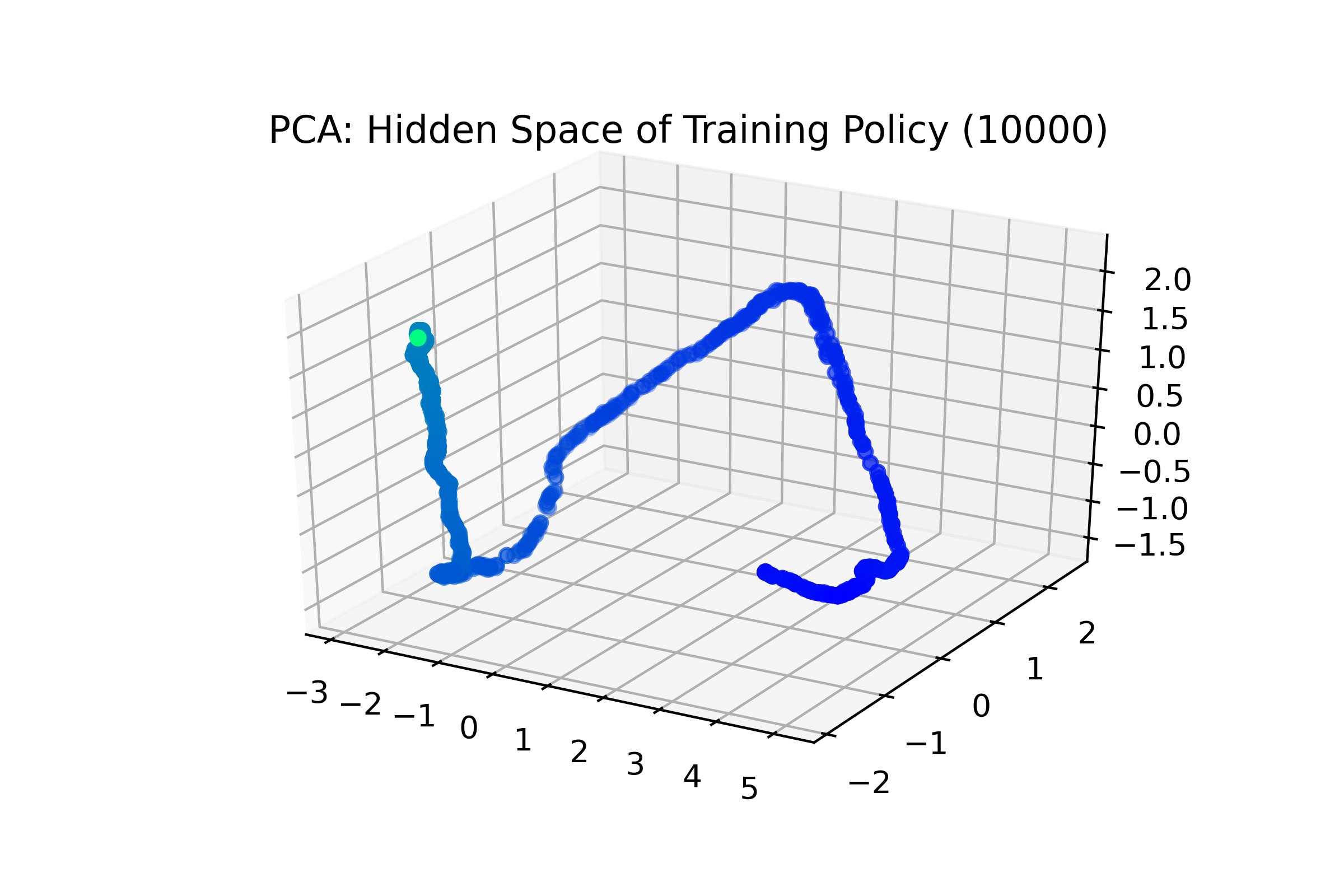}
        \includegraphics[width=.24\linewidth]{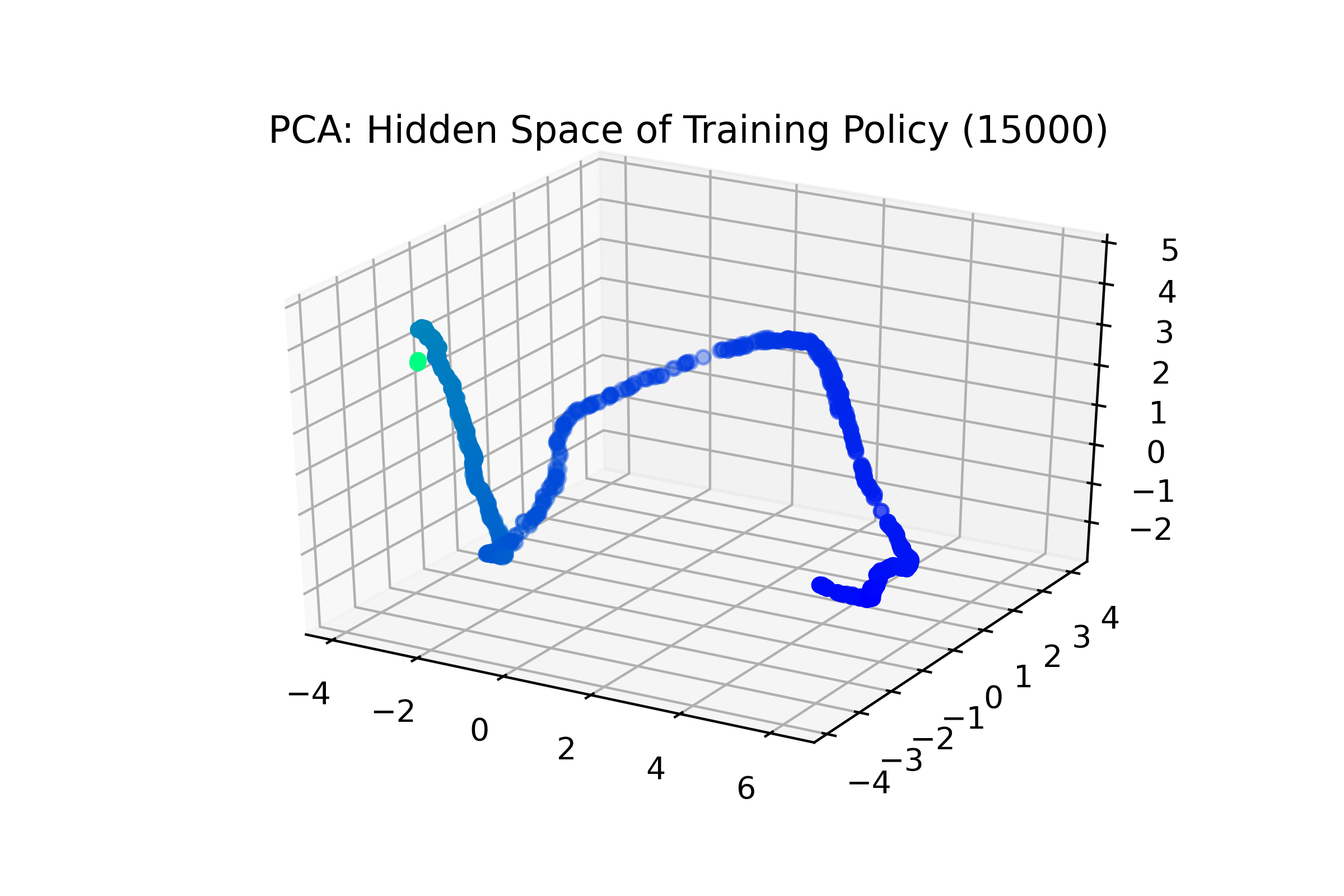}
    \end{subfigure}
    \caption{Image-based policy's internal representations of input states in the \textit{static, static} task at episode 0, 5k, 10k, and 15k of training (left to right).}
    \label{fig:pca-training-ss-img}
\end{figure}

One way to understand how these policy networks learn to organize their hidden representations is to visualize them throughout training. To do so, we take snapshots of the weights of policy networks at set intervals. The remaining plots in this section are arranged in the form of a timeline, read left to right, for each of the four experiment cases. Figure~\ref{fig:pca-training-ss-img}, below, shows the progress of an agent being trained on the full images of the environment in the \textit{static, static} task at episodes 0, 5,000, 10,000, and 15,000.  We note that, somewhere between episodes 5,000 and 10,000, the policy's internal representation becomes relatively more organized, as shown by the reduction in the plot's ``jitter''. This seems to coincide with the learning progress seen in Figure~\ref{fig:training-results}, which shows this particular agent lacking any real progress until after episode 5,000.

\begin{figure}[!h]
    \centering
    \begin{subfigure}{\textwidth}
    \centering
        \includegraphics[width=.24\linewidth]{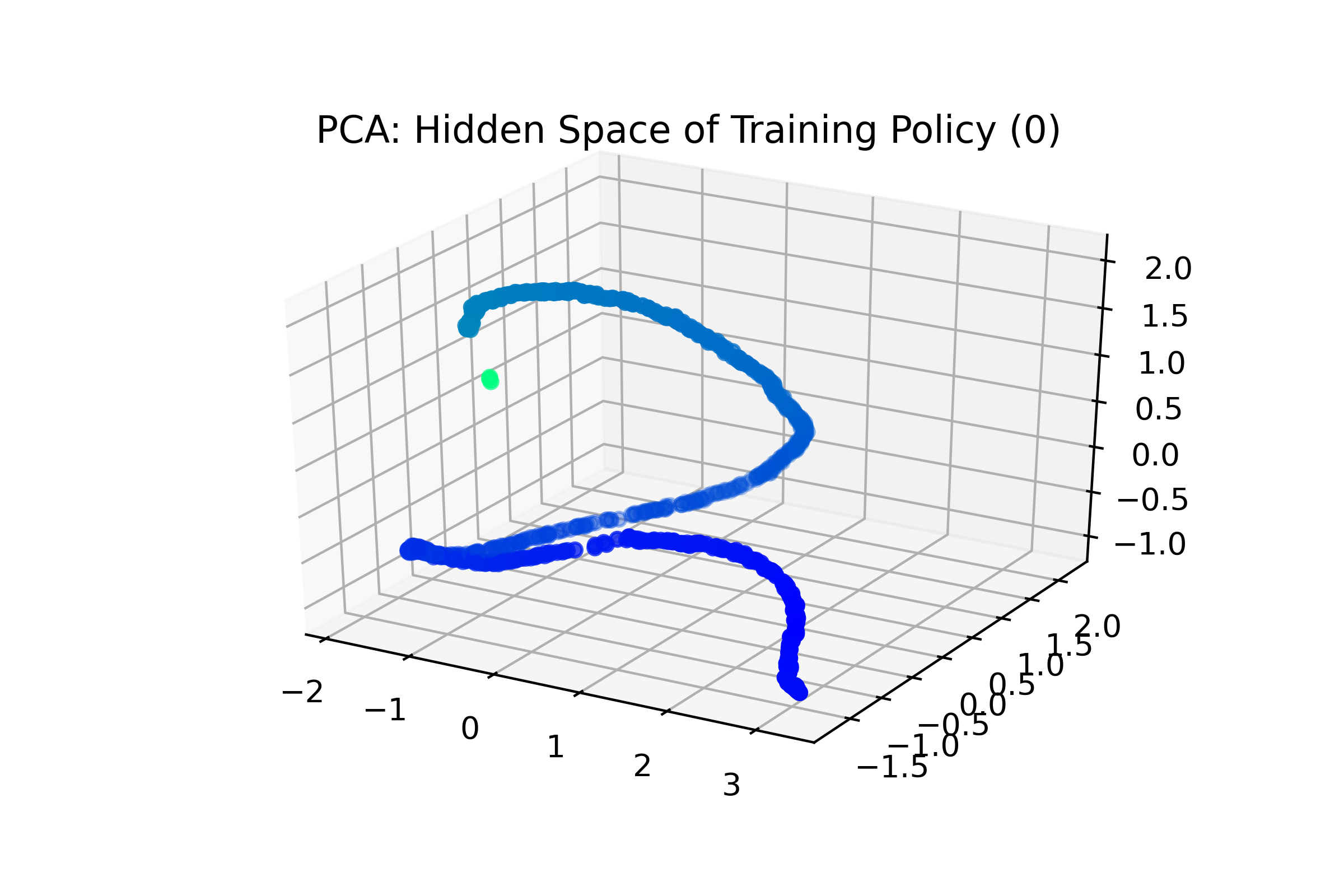}
        \includegraphics[width=.24\linewidth]{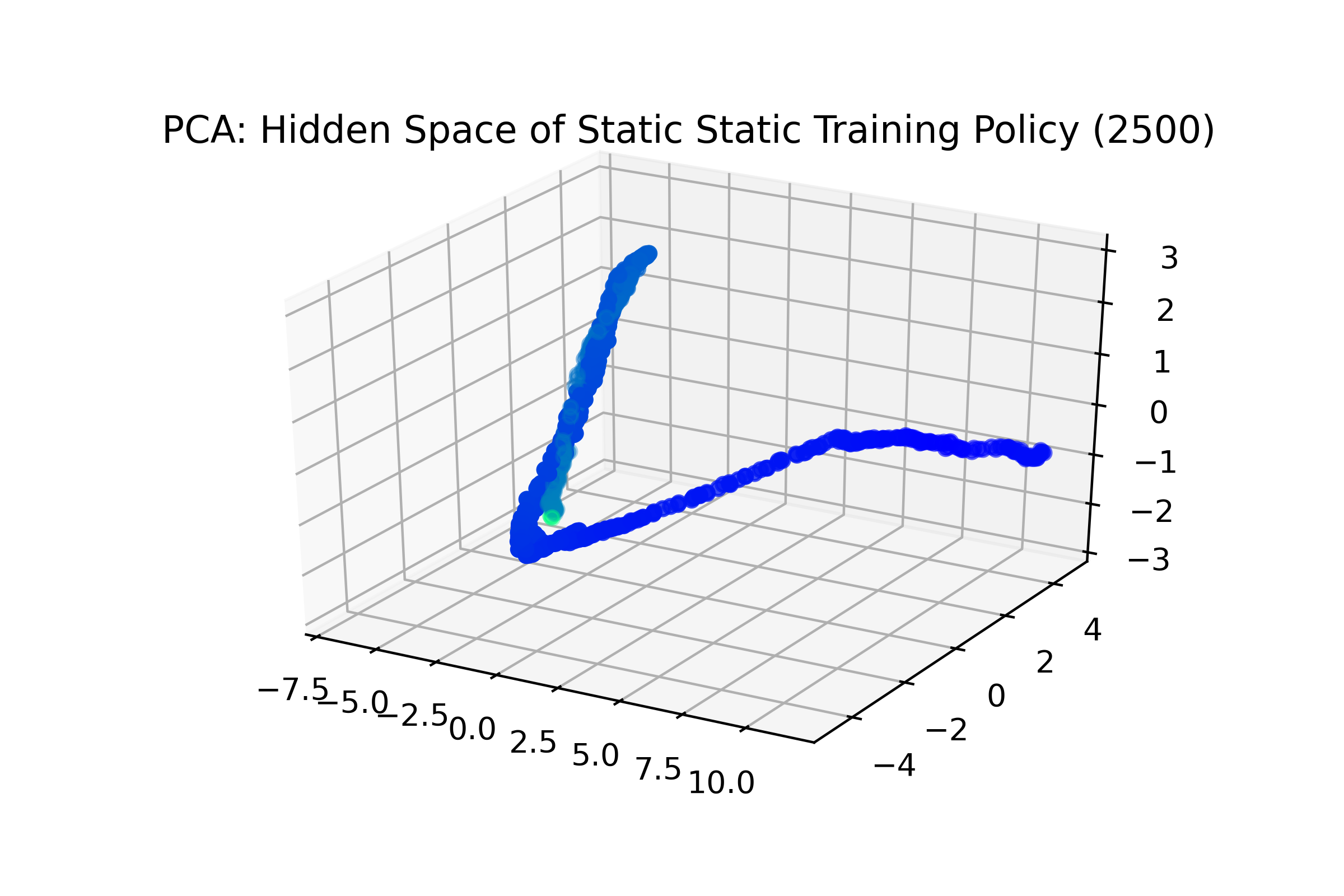}
        \includegraphics[width=.24\linewidth]{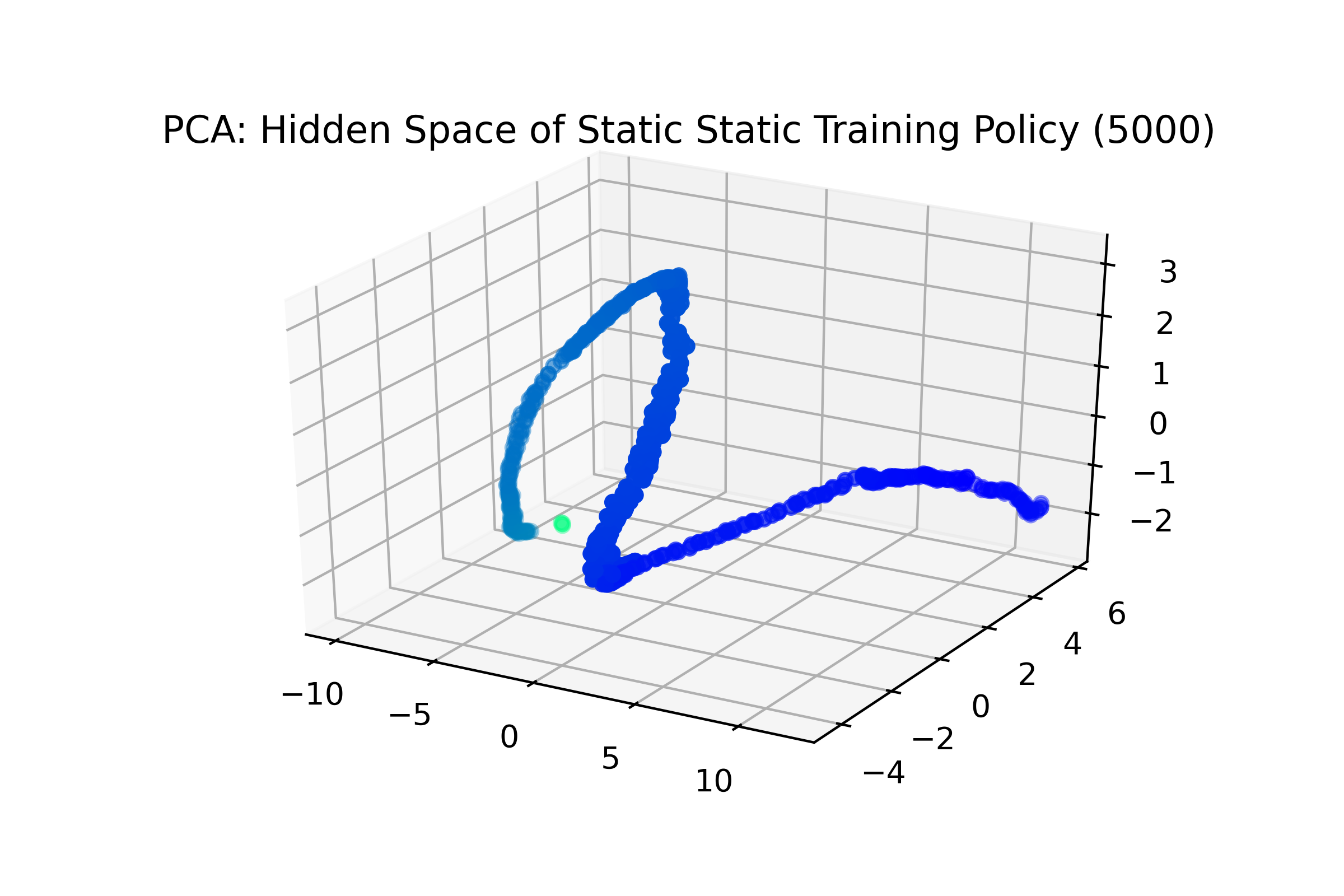}
        \includegraphics[width=.24\linewidth]{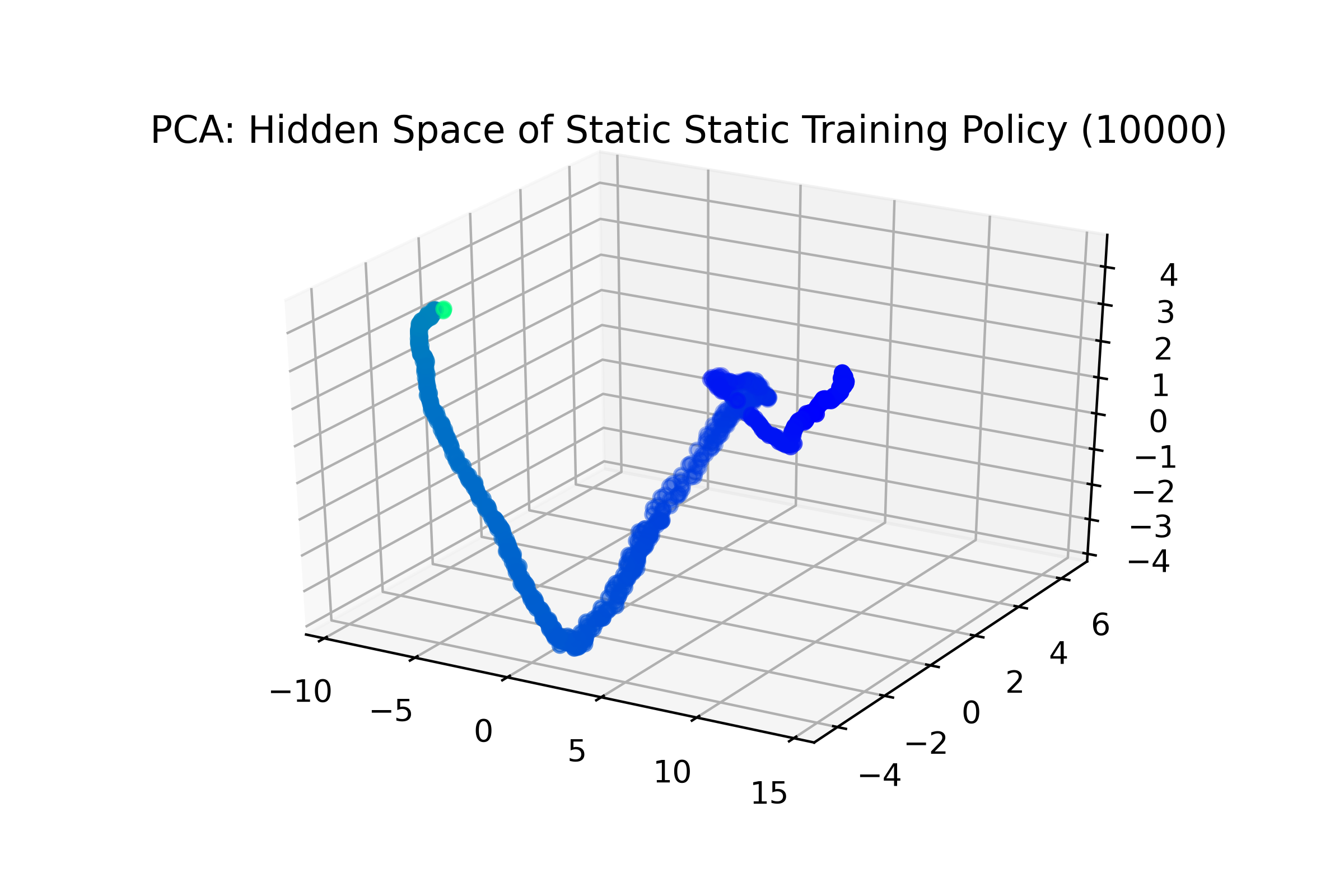}
    \end{subfigure}
    \caption{Latent-based policy's internal representations of input states in the \textit{static, static} task at episode 0, 2.5k, 5k, and 10k of training (left to right).}
    \label{fig:pca-training-ssl-img}
\end{figure}

Figure~\ref{fig:pca-training-ssl-img}, below, shows the same sort of timeline for the latent agent in the \textit{static, static} task at episodes 0, 2,500, 5,000, and 10,000. In contrast to the agent being trained on the full images, this agent's policy's internal representation is tightly organized from before training began. Coordinating this with the results in Figure~\ref{fig:training-results}, we highlight the fact that this agent begins making real progress in the very early stages of training.

\begin{figure}[!h]
    \centering
    \begin{subfigure}{\textwidth}
    \centering
        \includegraphics[width=.24\linewidth]{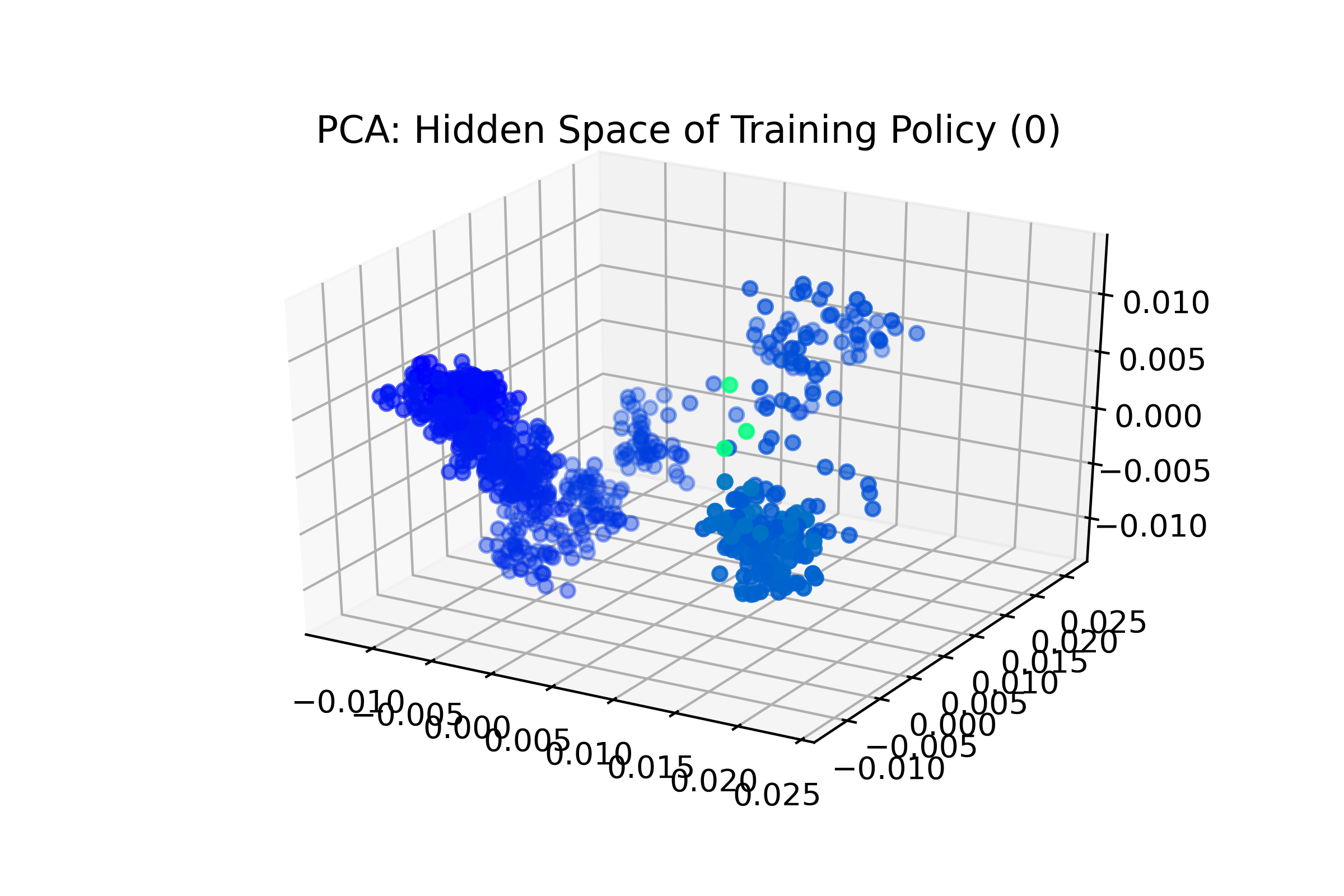}
        \includegraphics[width=.24\linewidth]{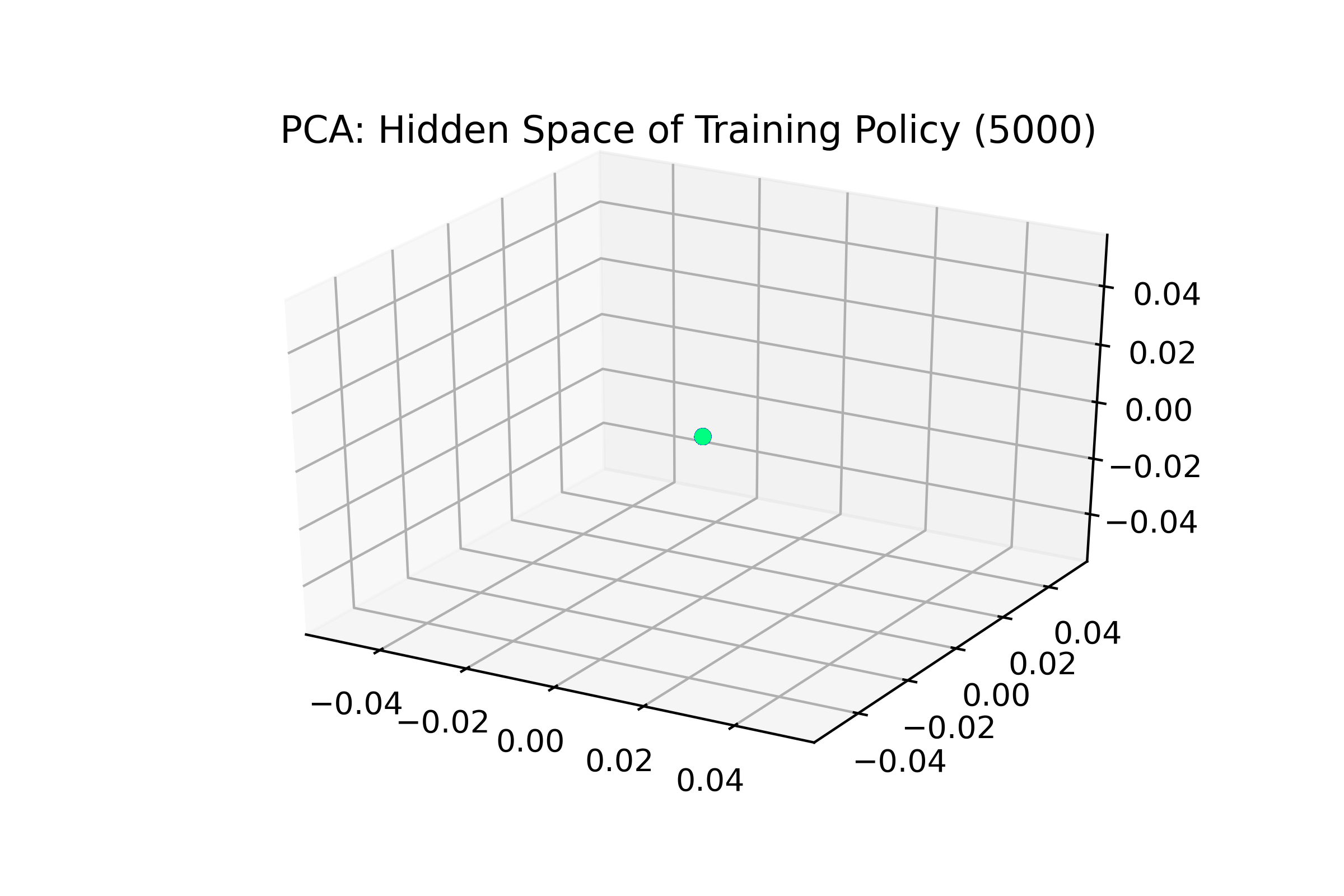}
        \includegraphics[width=.24\linewidth]{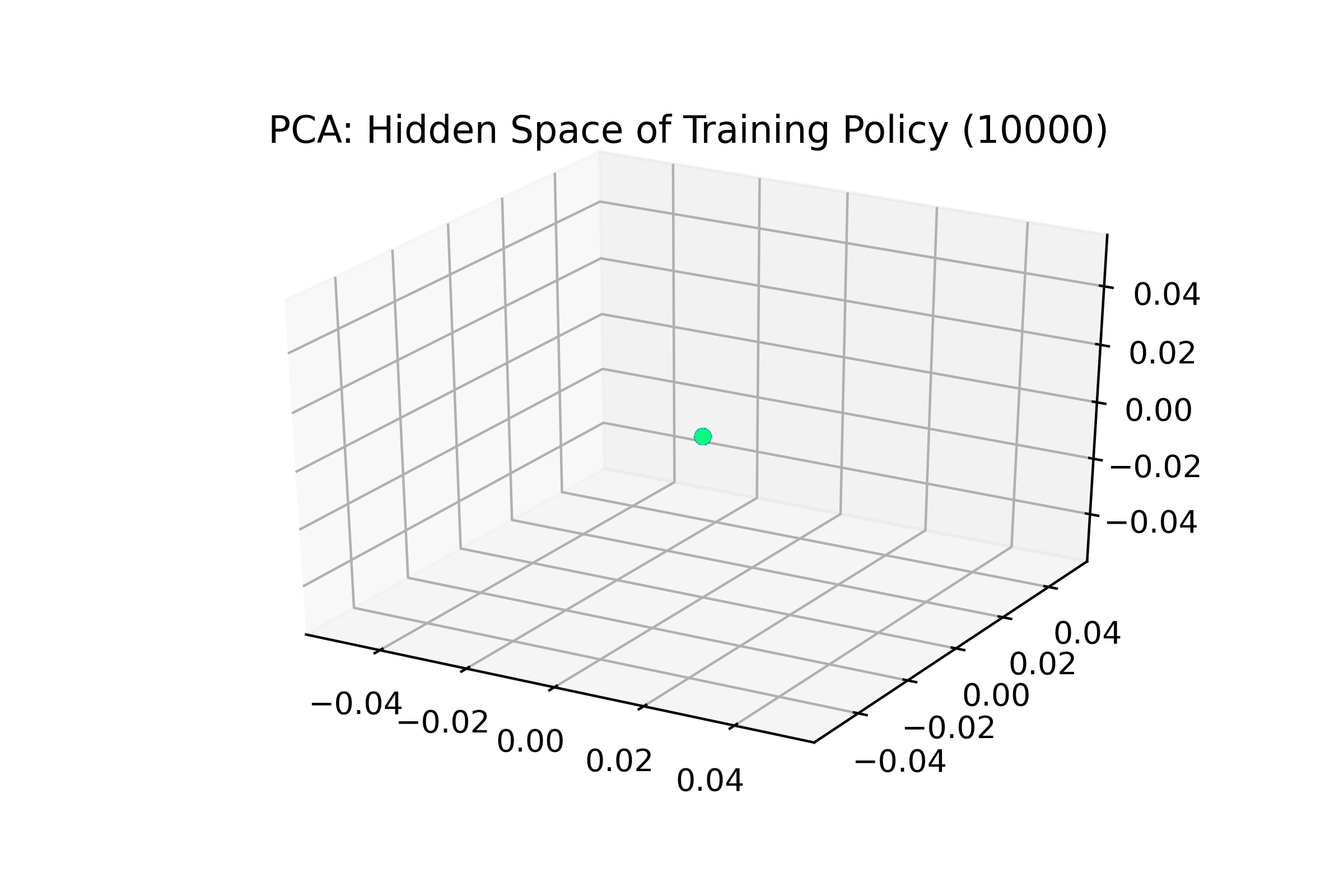}
        \includegraphics[width=.24\linewidth]{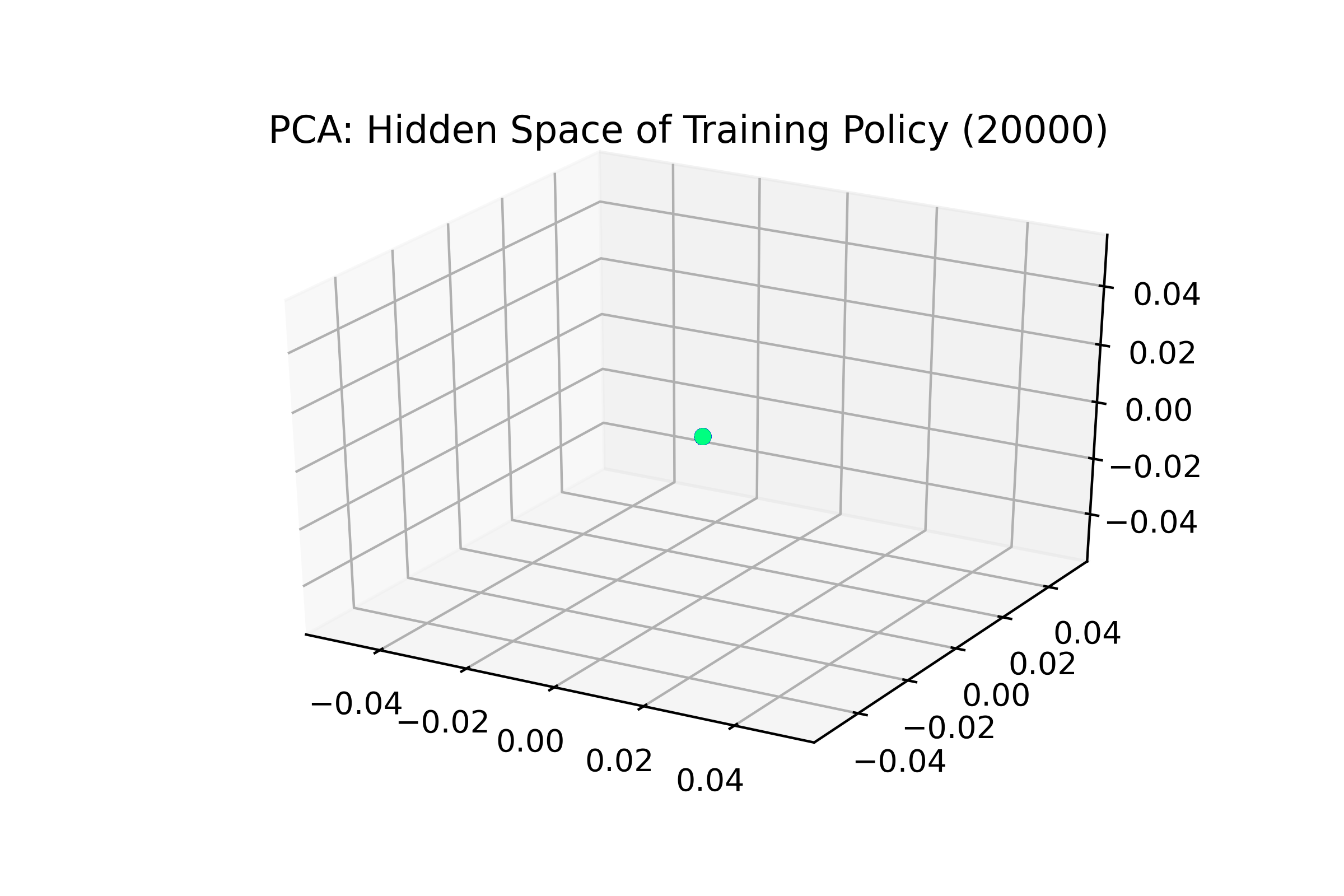}
    \end{subfigure}
    \caption{Image-based policy's internal representation collapsing on the \textit{static, random} task at episodes 0, 5k, 10k, and 20k of training (left to right).}
    \label{fig:pca-training-collapse}
\end{figure}

Figure~\ref{fig:pca-training-collapse}, below, displays the evolution of the hidden representations for an agent being trained on the full images in the \textit{static, random} task at episodes 0, 5,000, 10,000, and 20,000. Most notably, these hidden representations collapse at some point between weight initialization and episode 5,000. This is represented in the plots by (seemingly) a single point, which is actually the result of every point being placed at the exact same location in three dimensional space. This means that this neural network has failed to learn to differentiate, at all, between any of its inputs. These results coordinate logically with Figure~\ref{fig:training-results}, as this policy-task combination made no real learning progress at any time during the 1,000,000 steps of training.

\begin{figure}[!h]
    \centering
    \begin{subfigure}{\textwidth}
    \centering
        \includegraphics[width=.24\linewidth]{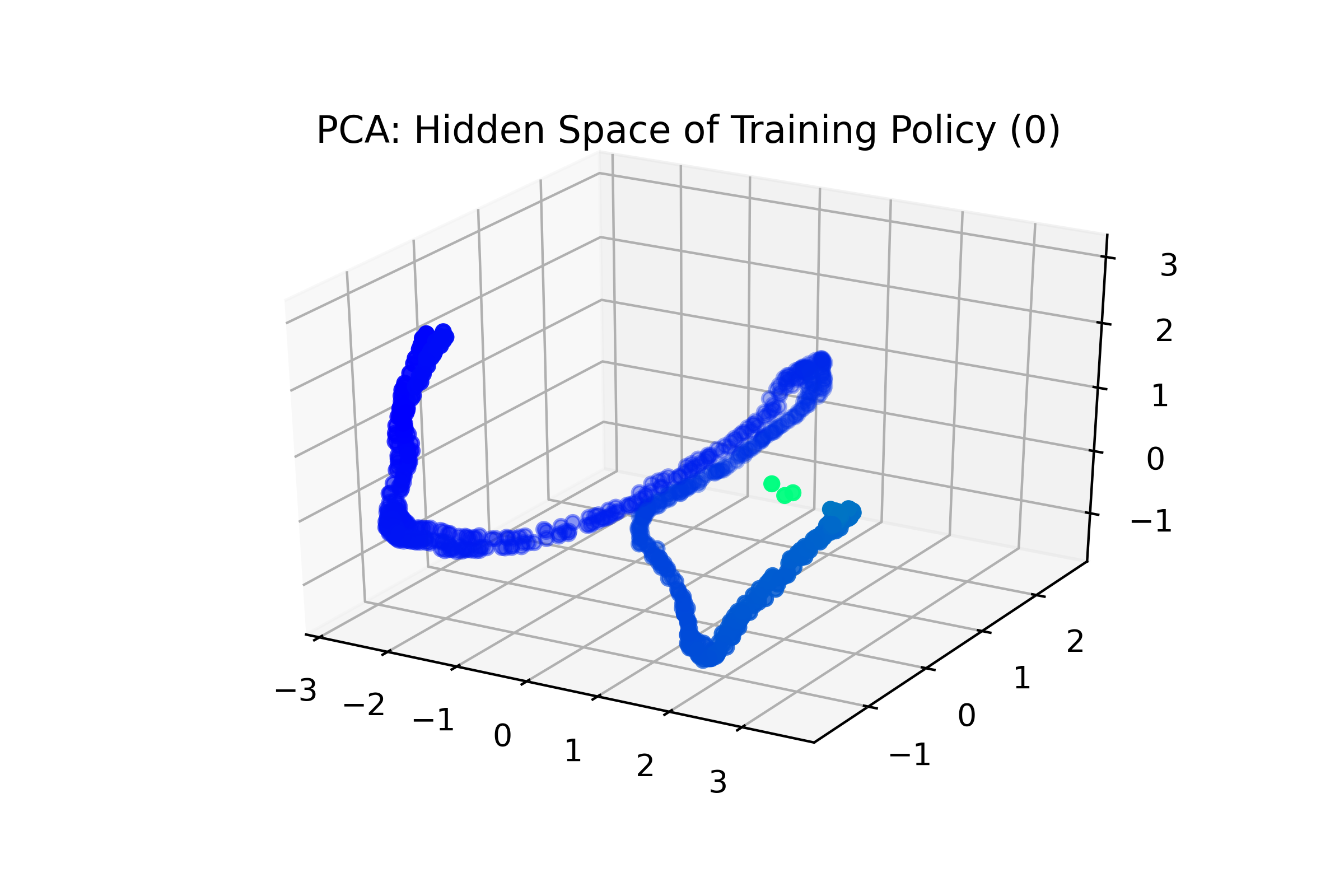}
        \includegraphics[width=.24\linewidth]{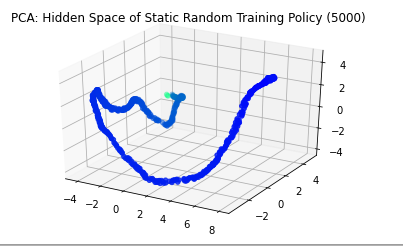}
        \includegraphics[width=.24\linewidth]{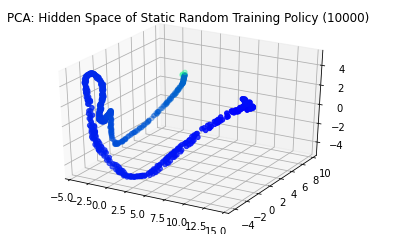}
        \includegraphics[width=.24\linewidth]{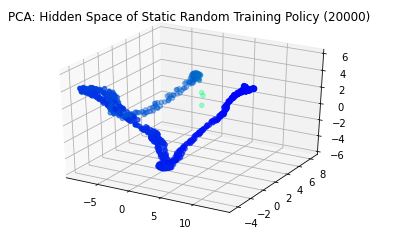}
    \end{subfigure}
    \caption{Latent-based policy's internal representations of input states in the \textit{static, static} task at episode 0, 2.5k, 5k, and 5k of training (left to right).}
    \label{fig:pca-training-srl-img}
\end{figure}

Finally, Figure~\ref{fig:pca-training-srl-img}, below, shows the progress of an agent being trained on latent representations in the \textit{static, random} task at episodes 0, 5,000, 10,000, and 20,000. Just as in the \textit{static, static} task, the latent representations seem to give the policy networks an innately organized hidden representation. However, when compared to the \textit{static, static} task, the organization of points is slightly noisier. This is most likely due to the increased stochasticity of the environment due to the object spawning in random locations every episode.

In total, these results suggest that a well-organized internal representation, in terms of the task's reward, is the first step in the learning progress of deep reinforcement learning agents. When we coordinate these temporal results with the overall plots of training progress, we see that the agents can only begin improving their action-selection by first being able to understand the important elements of their perceived environment. Finally, we observe the collapse of the internal representation of a convolutional network, resulting in the agent never improving at their task. This result suggests that a poor initial representation can have significant, negative impacts on the learning progress.



%% file: body/conclusion.tex
\section{Conclusion}
In this work, we train deep reinforcement learning agents on two versions of a robotic control task, one with an object that is placed in the same location every episode and one with an object that is placed in a random location every episode. Within each task, we train two different types of agents. One learns directly on the full images from the environment using a convolutional policy. The other learns on latent representations of the images that are produced by an auxiliary VAE by using a fully-connected policy. By independently learning a more-dense representation of the state space, the agents with the fully-connected policies are relieved of the burden of learning a good internal representation. As a result, they learn to complete both tasks more quickly.

Also, we analyze the hidden activations of the neural network policies by projecting them into three dimensional space via a linear transformation. In doing so, we reveal that the latent representations give the policy networks an internal representation that is innately organized around task rewards. In addition, we show that this phenomenon exists even under unusual weight initialization circumstances. Also, we observe the progress of hidden-representation organization for each type of policy-task combination. Our results show that a well-organized internal representation is a prerequisite to good policy learning. Finally, we observe a collapse phenomenon within a convolutional policy, suggesting that a poor initial representation can have catastrophic implications on the learning progress of deep reinforcement learning agents.